\documentclass[letterpaper, 10 pt, journal, twoside]{ieeetran}

\IEEEoverridecommandlockouts                              
\usepackage{graphicx}
\usepackage{caption}
\usepackage{epsfig}
\usepackage{amsmath,amssymb,enumerate}

\usepackage{amsthm}
\usepackage{setspace}
\usepackage[dvipsnames,svgnames,table]{xcolor}
\usepackage{mathrsfs}
\usepackage[mathscr]{euscript}
\usepackage{times}
\usepackage{svg}
\usepackage{stfloats}
\usepackage{url}
\usepackage{verbatim}
\usepackage{cite}
\usepackage{amssymb}
\usepackage{booktabs}
\usepackage{makecell}
\usepackage{threeparttable}
\definecolor{darkgreen}{rgb}{0,0.6,0}
\usepackage[bookmarks=true,colorlinks=true,pdfpagemode=UseNone,citecolor=darkgreen,linkcolor=black,urlcolor=BrickRed]{hyperref}
\usepackage{placeins}
\usepackage{subfigure}
\usepackage{bm}
\usepackage{multirow}

\usepackage[utf8]{inputenc}
\usepackage[T1]{fontenc}
\usepackage{textcomp}
\usepackage{arydshln}
\usepackage{balance}

\usepackage{hyperref}
\hypersetup{
    colorlinks=true,   
    linkcolor=blue,    
    urlcolor=blue     
}
\usepackage[
    top=57pt,
    bottom=43pt, 
    left=48pt,
    right=48pt, 
]{geometry}

\author{Zeyu Han$^{1}$, Shuocheng Yang$^{1}$, Minghan Zhu$^{2}$, Fang Zhang$^{3}$, Shaobing Xu$^{1}$, Maani Ghaffari$^{2}$, Jianqiang Wang$^{1}$
\thanks{Manuscript received: September 12, 2025; Revised December 9, 2025; Accepted January 9, 2026.}
\thanks{This paper was recommended for publication by Editor Sven Behnke upon evaluation of the Associate Editor and Reviewers’ comments.
This work was supported by the National Natural Science Foundation of China, the Key Project (52131201), and National Natural Science Foundation of China, Science Fund for Creative Research Groups (52221005). \textit{(Corresponding author: Fang Zhang.)}} 
\thanks{$^{1}$ Z. Han, S. Yang, S. Xu, and J. Wang are with the School of Vehicle and Mobility, Tsinghua University, Beijing, China.
        {\tt\footnotesize \{hanzy21, ysc24\}@mails.tsinghua.edu.cn},  {\tt\footnotesize \{shaobxu, wjqlws\}@mail.tsinghua.edu.cn}}%
\thanks{$^{2}$ M. Zhu and M. Ghaffari are with the Computational Autonomy and Robotic Laboratory, University of Michigan, Ann Arbor, USA.
        {\tt\footnotesize \{minghanz, maanigj\}@umich.edu}}%
\thanks{$^{3}$ F. Zhang is with the State Key Laboratory of Intelligent Green Vehicle and Mobility, Tsinghua University, Beijing, China.
        {\tt\footnotesize zhang\_fang@mail.tsinghua.edu.cn}}%
\thanks{Digital Object Identifier (DOI): see top of this page.}
}

\title{Equi-RO: A 4D mmWave Radar Odometry via Equivariant Networks}

\begin{document}

\maketitle

\begin{abstract}
Autonomous vehicles and robots rely on accurate odometry estimation in GPS-denied environments. While LiDARs and cameras struggle under extreme weather, 4D mmWave radar emerges as a robust alternative with all-weather operability and velocity measurement. In this paper, we introduce Equi-RO, an equivariant network-based framework for 4D radar odometry. Our algorithm pre-processes Doppler velocity into invariant node and edge features in the graph, and employs separate networks for equivariant and invariant feature processing. A graph-based architecture enhances feature aggregation in sparse radar data, improving inter-frame correspondence. Experiments on an open-source dataset and a self-collected dataset show Equi-RO outperforms state-of-the-art algorithms in accuracy and robustness. Overall, our method achieves 10.7\% and 13.4\% relative improvements in translation and rotation accuracy, respectively, compared to the best baseline on the open-source dataset. 
\end{abstract}


\section{Introduction} \label{Sec:Introduction}

\IEEEPARstart{O}{dometry} is indispensable for autonomous vehicles and robots, supporting downstream perception, planning, and control. When Global Positioning System (GPS) signals are degraded, localization relies primarily on odometry from onboard sensors such as LiDARs and cameras. However, these sensors often struggle in extreme weather, whereas 4D millimeter-wave (mmWave) radar (4D radar) provides compact, cost-efficient, all-weather, velocity-measuring, and long-range sensing, making it a promising alternative for robust perception and localization under adverse conditions~\cite{han20234d}. Current 4D radar odometry mostly adapts traditional LiDAR methods and exploits Doppler velocity and Radar Cross Section (RCS) for ego-velocity estimation and point cloud registration~\cite{zhang20234dradarslam, zhuang20234d}, while a few learning-based works employ convolutional and recurrent neural networks for 4D radar~\cite{lu2020milliego} and radar-camera fusion~\cite{zhuo20234drvo} odometry.

\begin{figure}
    \centering
    \includegraphics[width = 0.9\linewidth]{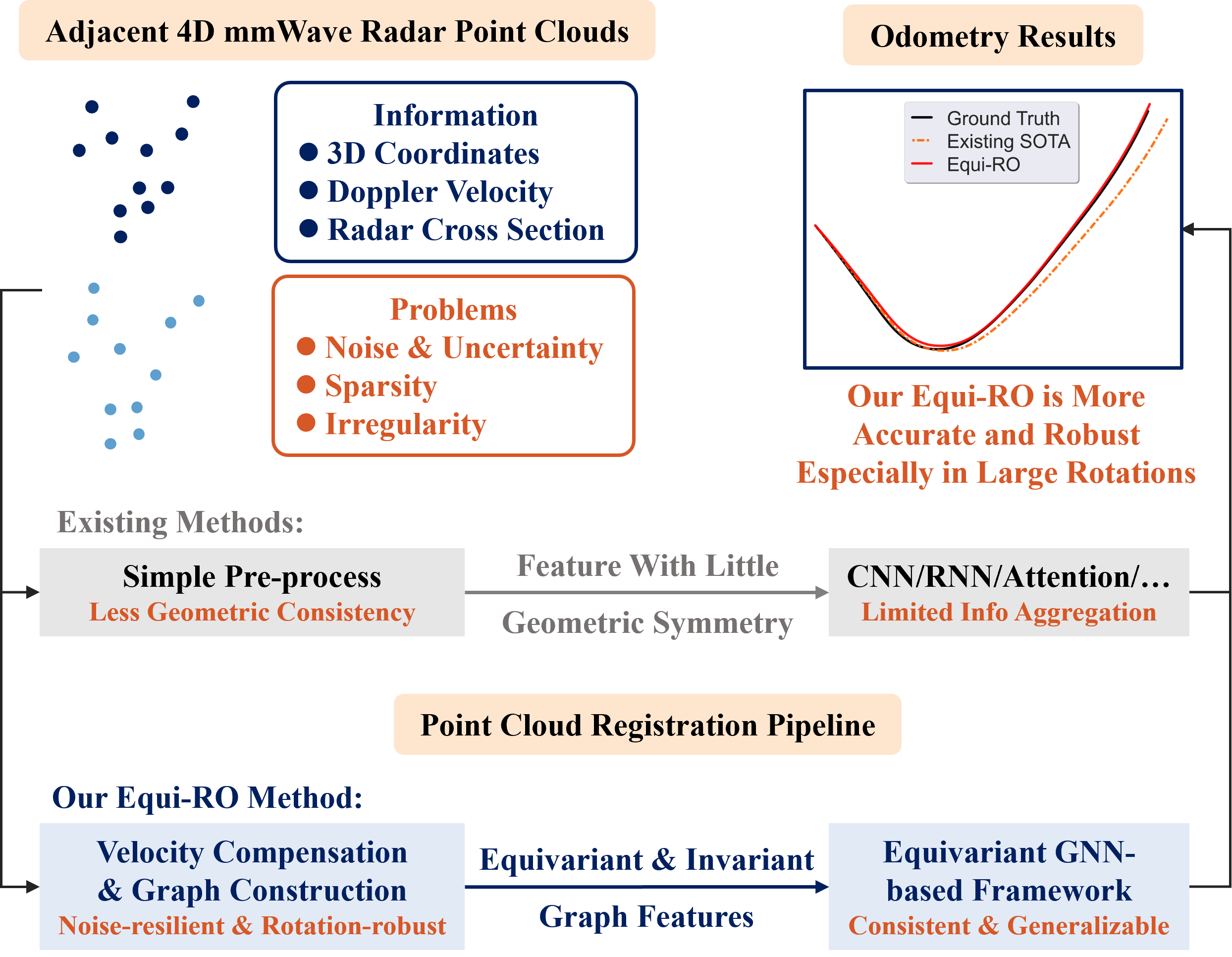}
    \caption{Our Equi-RO method incorporates Doppler velocity compensation and graph construction to preserve noise-resilient and rotation-robust features. The derived equivariant and invariant features are fed into an equivariant Graph Neural Network (GNN)-based framework for consistent and generalizable point cloud registration, yielding more accurate odometry results, especially in large rotations. Trajectory comparison on a partial segment of the \textit{loop3} split from the NTU4DRadLM dataset~\cite{zhang2023ntu4dradlm}, where both methods are initialized to the ground truth at the segment's start, comparing against~\cite{zhang20234dradarslam}.}
    \label{Fig:first_overview}
\vspace{-5mm}
\end{figure}

In learning-based odometry, feature extraction and matching are critical. However, features in adjacent frames may become highly inconsistent when the ego-vehicle moves rapidly or turns sharply. Equivariant neural networks have emerged as a promising approach to address this issue~\cite{cohen2016group, zhu2023e2pn}. They learn features that transform predictably under, e.g., rigid body transformation group SE(3) (equivariance property), or remain fixed under such transformations (invariance property).

In 4D radar sensing, the Doppler effect enables estimation of targets' radial relative velocity. However, Doppler velocity is neither equivariant nor invariant under the sensor's transformations, as it depends on both the sensor's and target's motion, limiting the applicability of equivariant networks to 4D radar odometry.
Moreover, 4D radar point clouds, with typically only thousands of points per frame, are noisier, sparser, and more irregular than LiDAR point clouds. Consequently, effectively aggregating point-wise features from spatial neighbors remains a challenge for learning-based 4D radar odometry. 

In this paper, we propose \textbf{Equi-RO}, a novel 4D radar odometry framework built upon equivariant networks (Fig.~\ref{Fig:first_overview}). We design a pre-processing pipeline to derive invariant Doppler features and construct a graph from the point cloud to mitigate noise and sparsity. Compared with raw point sets or voxel grids, a graph representation explicitly connects spatially and kinematically consistent neighbors, which can suppress isolated noise and stabilize feature extraction for 4D radar point clouds.
The invariant velocity features and equivariant network design enable our method to remain robust in challenging real-world driving scenarios, where existing methods often degrade. Leveraging both a public dataset~\cite{zhang2023ntu4dradlm} and a self-collected dataset, we demonstrate consistent improvements over state-of-the-art 4D radar odometry approaches.

In summary, our main contributions are as follows:

\begin{enumerate}[1.]

    \item We develop a novel 4D radar odometry algorithm based on equivariant neural networks, providing accurate and robust odometry estimates under large rotations, as commonly encountered in sharp turns and challenging driving maneuvers.
    \item We design a dedicated pre-processing pipeline for Doppler velocity that derives invariant features, thereby stabilizing motion-related features. We also design a framework that separately extracts equivariant and invariant features to preserve geometric consistency, making it applicable to 4D radar point clouds with physical attributes.
    \item We conduct extensive experiments on both an open-source dataset and a self-collected dataset, along with ablation studies, to validate the effectiveness of the proposed method across diverse scenarios. The project is available at: \url{https://github.com/hanzy21/Equi-RO}.
\end{enumerate}

\section{Related Works}  \label{Sec:Related}
In this section, we briefly review existing algorithms for 3D point cloud odometry and 4D radar odometry, followed by a discussion of equivariant neural networks.

\subsection{3D Point Cloud Odometry}
3D point cloud odometry for LiDAR and RGB-D cameras can be categorized into traditional and learning-based methods. Traditional methods mainly utilize geometric registration, such as Iterative Closest Point (ICP)~\cite{besl1992method} and Normal Distributions Transform (NDT)~\cite{biberNormalDistributionsTransform2003}. Subsequent Generalized-ICP (GICP)~\cite{segalGeneralizedicp2009} and KISS-ICP~\cite{vizzo2023kiss} further improve robustness through local surface modeling. Learning-based methods such as FCGF~\cite{choy2019fully} and DeepVCP~\cite{lu2019deepvcp} replace handcrafted features with learned correspondence models on dense point clouds.

Despite their success on dense point clouds, these methods' heavy reliance on dense and smooth geometry limits their applicability to sparse and noisy data from 4D radars.

\subsection{4D Radar Odometry}
Several core ideas of 3D point cloud odometry including geometric registration and local feature extraction are transferable to 4D radar odometry. However, leveraging radar-specific measurements (Doppler and RCS) and handling the sparsity and noise of radar point clouds remain open challenges. These issues have led to recent works on 4D radar odometry.

For measurements utilization, Doppler velocity is commonly employed for ego-velocity estimation~\cite{doer2020ekf, wang2024riv}, dynamic object removal~\cite{zhuang20234d, herraez2024radar} and velocity-aware attention module design~\cite{lu2023efficient, zhuo20234drvo}, while RCS is often used for point cloud filtering~\cite{huang2024less, herraez2025ground} and weighted point matching~\cite{kim2024radar4motion} since it indicates the intensity of the point.

As for handling the sparsity and noise of radar point clouds, classical 4D radar odometry uses scan-to-submap~\cite{zhuang20234d, kung2021normal}, adaptive probability-based~\cite{zhang20234dradarslam, li20234d}, and voxel-based~\cite{seok2025radar4voxmap} registration to improve the robustness. For learning-based methods, RaFlow~\cite{ding2022self} predicts scene flow, while CAO-RONet~\cite{li2025cao} incorporates local completion to match noisy points.

To the best of our knowledge, although GNNs can explicitly aggregate features from spatially neighboring points, they have not yet been applied to 4D radar odometry.

\subsection{Equivariant Neural Networks}
Equivariant neural networks produce predictable outputs under specific input transformations, enhancing generalizability and interpretability. Since Cohen and Welling introduced group equivariance into neural networks~\cite{cohen2016group}, various equivariant designs have been proposed for GNNs~\cite{du2022se, satorras2021n}. For example, Equivariant Graph Neural Network (EGNN)~\cite{satorras2021n} maintains equivariance to the Euclidean group E(n), which includes translations, rotations, and reflections, by incorporating the relative distances in the edge feature aggregation.

For point cloud analysis, Tensor Field Network~\cite{thomas2018tensor} achieves SE(3)-equivariance via spherical harmonics, while Vector Neurons (VN)~\cite{deng2021vector} and Lie Neurons (LN)~\cite{lin2024lie} extend traditional Multi-Layer Perceptrons (MLPs) to vector features, achieving equivariance to 3D rotation group SO(3) and semi-simple Lie groups, respectively. EPN~\cite{chen2021equivariant} and E2PN~\cite{zhu2023e2pn} further introduce SE(3)-equivariant convolutional architectures for 3D point clouds. However, equivariant models remain largely underexplored for 4D radar point clouds.

\section{Methodology}  \label{Sec:Methodology}
This section provides a detailed description of our proposed Equi-RO algorithm. 

\subsection{Overview}\label{Subsec:Overview}

\begin{figure*}
    \centering
    \includegraphics[width = 0.88\linewidth]{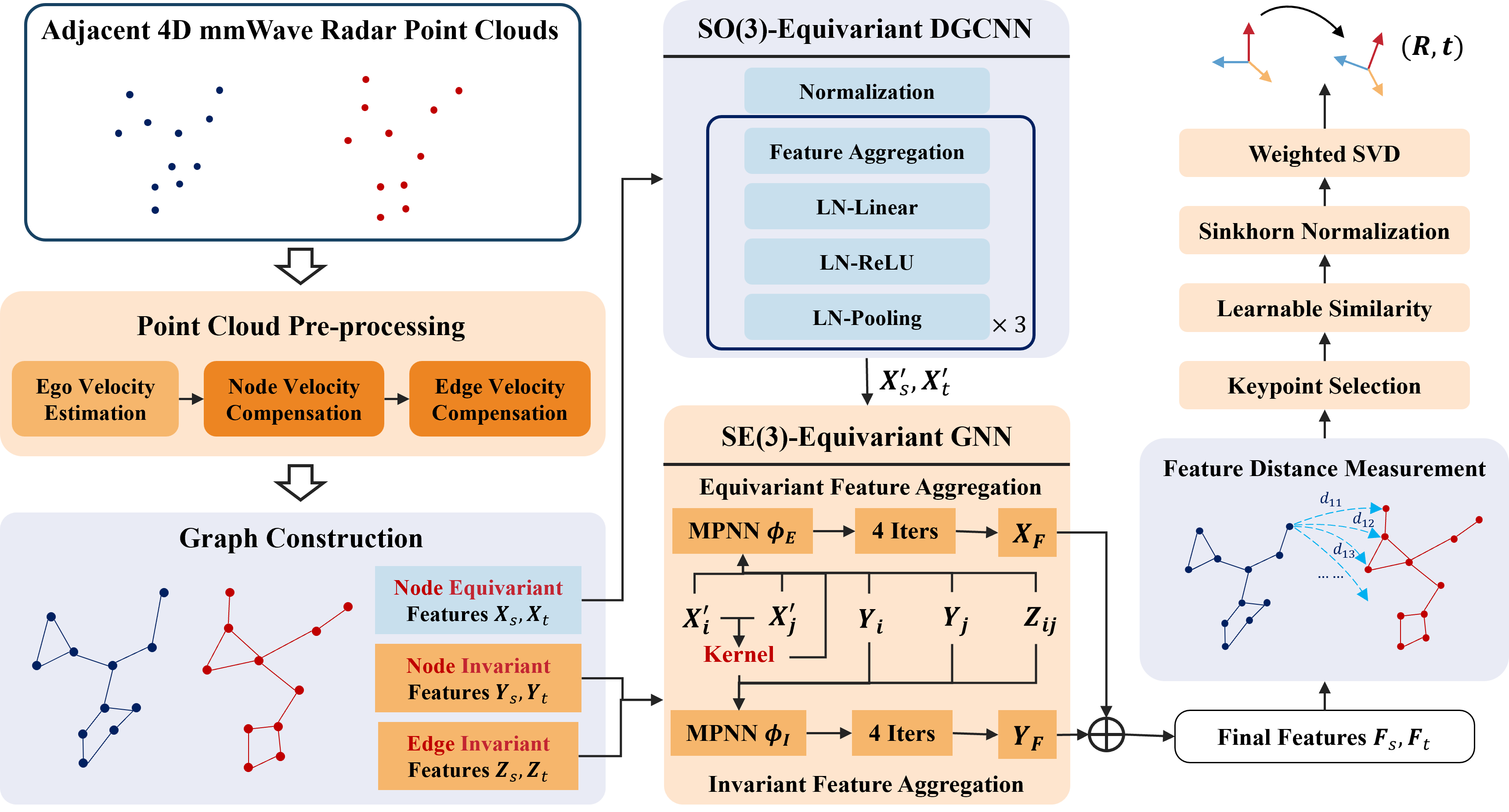}
    \caption{Overview of the Equi-RO algorithm. Doppler velocity is used to derive invariant node and edge features, which are combined with node and edge features within a unified graph-based network framework to robustly estimate the relative transformation $(\bm{R}, \bm{t})$ between two consecutive radar frames.}
    \label{Fig:overview}
\vspace{-5mm}
\end{figure*}

Fig.~\ref{Fig:overview} shows the pipeline of Equi-RO. Standard 4D radar odometry faces two primary challenges: the sparsity of radar point clouds and the ego-motion dependent nature of Doppler velocities. To address sparsity and noise, we structure the point cloud as a graph rather than isolated points. This allows us to explicitly define edge features that capture local relative patterns, aggregating information from neighbors to improve robustness. To address the motion coupling of Doppler velocity, we mathematically derive invariant compensated velocities to decouple the object's motion from the ego-vehicle's rotation.

Based on these pre-processed inputs, we extract equivariant geometric structures (e.g., 3D coordinates that transform consistently with the sensor) and invariant physical features (e.g., RCS and our derived Doppler velocity that remain unchanged under the sensor's rotation or translation), and process these features with different networks. The equivariant features enable the network to reason about relative motion during scan matching, while the invariant features remain consistent across different sensor orientations, providing pose-independent cues. Specifically, an SO(3)-equivariant network based on LN~\cite{lin2024lie} is employed to extract node-level equivariant features, while both node- and edge-level invariant features are fed into an EGNN-based~\cite{satorras2021n} module. The output features are used to estimate the relative transformation $(\bm{R}, \bm{t})$ between the two input frames, achieving robust registration even under large rotations.

\subsection{Point Cloud Pre-processing}\label{Subsec:Preprocessing}
As mentioned above, 4D radar can directly measure the radial relative velocity of targets via the Doppler effect. In this section, we fully exploit the Doppler velocity to estimate the ego-velocity and to derive compensated node and edge velocities for subsequent GNN-based equivariant framework.

\subsubsection{Ego-velocity Estimation}
We apply an iterative reweighted least squares algorithm~\cite{yang2020graduated} to estimate the ego-velocity $\mathbf{v}_{ego}$. For a point $i$ with absolute velocity $\mathbf{v}_{i}^{abs}$, its Doppler velocity $v_{i}^{dop}$ can be represented as:
\begin{equation}
    v_{i}^{dop}=\frac{\bm{\rho}_i^T}{\left\|\bm{\rho}_i\right\|} \cdot (\mathbf{v}_{ego} - \mathbf{v}_{i}^{abs}), 
\end{equation}
where $\bm{\rho}_i$ is the coordinate of this point. We define the Doppler residual $v_i^r = \left\| v_{i}^{dop} - \frac{\bm{\rho}_i^T}{\left\|\bm{\rho}_i\right\|} \cdot \mathbf{v}_{ego} \right\|$, which should be zero for stationary points.  To robustly suppress moving points, we can formulate the following weighted least squares problem for a point cloud with $N$ points:
\begin{equation}
    \underset{\mathbf{v}_{ego}}{\min} \sum_{i=1}^{N} \omega_i {v_i^r}^2.
\end{equation}
The weight $\omega_i$ indicates the likelihood of a point being stationary and is iteratively updated as $\omega_i = 1 / (v_i^r + \epsilon)$, with $\epsilon = 1 \times 10^{-5}$. Upon convergence, $\mathbf{v}_{ego}$ yields a robust ego-velocity estimate.

\subsubsection{Node Velocity Compensation}
Once $\mathbf{v}_{ego}$ is estimated, the compensated node velocity $v_{i}^{dop'}$ for each point $i$ is:
\begin{equation}
    v_{i}^{dop^{'}} = v_{i}^{dop} - \frac{\bm{\rho}_i^T}{\left\|\bm{\rho}_i\right\|} \cdot \mathbf{v}_{ego} = -\frac{\bm{\rho}_i^T}{\left\|\bm{\rho}_i\right\|} \cdot \mathbf{v}_{i}^{abs}.\label{n1}
\end{equation} 
As illustrated in Fig.~\ref{Fig:coms} (left), $v_{i}^{dop'}$ is the radial component of the absolute velocity $\mathbf{v}_{i}^{abs}$ of point $i$. If the radar coordinate frame is rotated by a matrix $\bm{R} \in SO(3)$ with $\bm{R}^{T}\bm{R}=\bm{I}$, $v_{i}^{dop'}$ remains the same as:

\begin{align}
 v_{i}^{dop^{''}} &= -\frac{(\bm{R\rho}_i)^T}{\left\|\bm{R\rho}_i\right\|} \cdot \bm{R}\mathbf{v}_{i}^{abs}=-\frac{\bm{\rho}_i^T\bm{R}^T\bm{R}\mathbf{v}_{i}^{abs}}{\left\|\bm{\rho}_i\right\|}  \nonumber \\
  &= -\frac{\bm{\rho}_i^T}{\left\|\bm{\rho}_i\right\|} \cdot \mathbf{v}_{i}^{abs} = v_{i}^{dop^{'}}.
  \vspace{-5mm}
\end{align}

Therefore, $v_{i}^{dop'}$ is invariant under SO(3) transformations of the radar, providing a stable feature representation for point cloud registration under large rotational variations.

\subsubsection{Edge Velocity Compensation}
A graph is constructed from the radar point cloud, so it is necessary to define an edge feature based on Doppler velocity. The compensated edge relative velocity $v_{ij}$ between points $i$ and $j$ is formulated as:
\begin{equation}
    v_{ij} = \frac{v_{i}^{dop^{'}}\left\|\bm{\rho}_i\right\|-v_{j}^{dop^{'}}\left\|\bm{\rho}_j\right\|}{\left\|\bm{\rho}_i-\bm{\rho}_j\right\|}. \label{e1}
\end{equation}
Since both $v_{i}^{dop'}$ and $v_{j}^{dop'}$ are individually SO(3)-invariant, and rotations preserve vector norms and pairwise distances, it is evident that $v_{ij}$ is also SO(3)-invariant. 

\begin{figure}[t]
    \centering
    \subfigure{
        \includegraphics[width=0.2\textwidth]{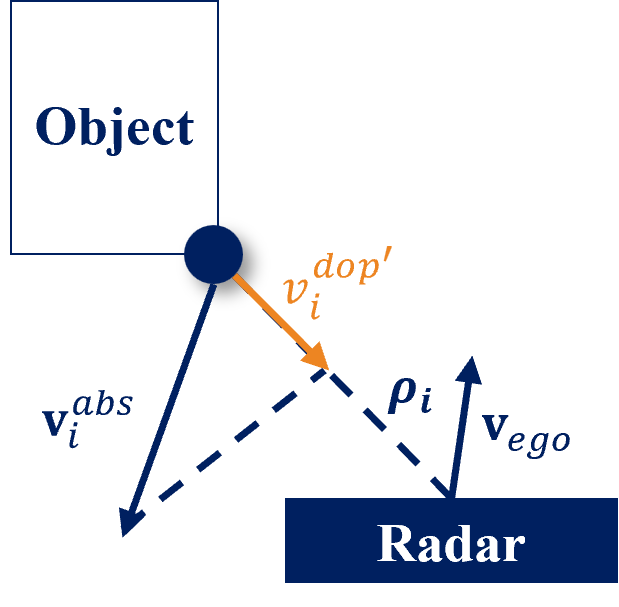}
        \label{Fig:node_com}}
        \subfigure{
        \includegraphics[width=0.2\textwidth]{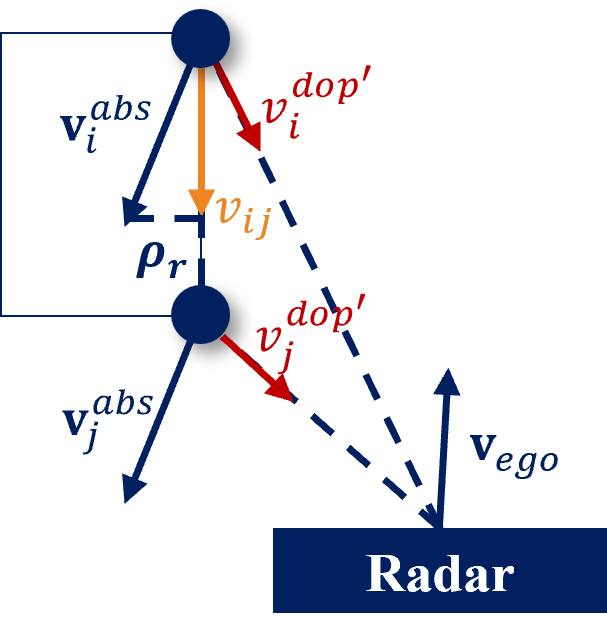}
        \label{Fig:edge_com}}
    \caption{Illustrations of (left) node and (right) edge velocity compensation. The yellow arrow in the left figure indicates the compensated node velocity, which remains SO(3)-invariant. The yellow arrow in the right figure is the compensated edge velocity when the two nodes share the same absolute velocity, which is SE(3)-invariant.}
    \label{Fig:coms}
    \vspace{-4mm}
\end{figure}

Moreover, as Fig.~\ref{Fig:coms} (right) presents, if both nodes of an edge have identical absolute velocities, i.e., $\mathbf{v}_{i}^{abs}=\mathbf{v}_{j}^{abs}$, we can substitute Eq. \eqref{n1} into Eq. \eqref{e1}, $v_{ij}$ as:
\begin{equation}
    v_{ij} = \frac{(\bm{\rho}_j^{T}-\bm{\rho}_i^{T})}{\left\|\bm{\rho}_i-\bm{\rho}_j\right\|}   \cdot \mathbf{v}_{i}^{abs}=\frac{\bm{\rho}_r^{T}}{\left\|\bm{\rho}_r\right\|} \cdot \mathbf{v}_{i}^{abs},
\end{equation}
where $\bm{\rho}_r$ denotes the relative coordinates from $i$ to $j$. If the radar undergoes a rigid-body transformation $(\bm{R}, \bm{t}) \in SE(3)$, the translation $\bm{t}$ does not change $\mathbf{v}_{i}^{abs}$ and $\bm{\rho_{r}}$. So we have:

\begin{align}
     v_{ij}^{'}  &= \frac{(\bm{R\rho}_r)^{T}}{\left\|\bm{R\rho}_r\right\|} \cdot (\bm{R}\mathbf{v}_{i}^{abs}) =  \frac{\bm{\rho}_r^{T}\bm{R}^T \bm{R}\mathbf{v}_{i}^{abs}}{\left\|\bm{\rho}_r\right\|} \\ \nonumber
     & = \frac{\bm{\rho}_r^{T}}{\left\|\bm{\rho}_r\right\|} \cdot \mathbf{v}_{i}^{abs} = v_{ij}.
\end{align}

Hence, the edge compensated velocity $v_{ij}$ is SE(3)-invariant when its two nodes share the same velocity, which holds when they belong to a rigid object with no angular velocity.

\subsection{Graph Construction}\label{Subsec:Graph}
After pre-processing, the graph is constructed for each frame of point cloud by treating each point as a node. A cost function $f(i,j)$ is defined to select node pairs for edge generation, preferring neighboring points with similar velocities:
\begin{equation}
    f(i,j) = \left\| \bm{\rho}_i - \bm{\rho}_j\right\| + \lambda \left\| v_{i}^{dop^{'}} - v_{j}^{dop^{'}}\right\|,
\end{equation}
where $\lambda$ is a weighting parameter. Edges are established between node $i$ and its $k$ nearest neighbors in the feature space.

Subsequently, to effectively fuse spatial and physical information, we organize the input into three distinct sets based on their transformation properties: 
\par \noindent   \textbf{Geometric equivariant node feature set} $\mathbf{X}\in \mathbb{R}^{N\times 3}$: This set contains Euclidean coordinates $\bm{\rho}_i$ of each node $i$ which reflect the geometric information and are inherently SE(3)-equivariant:
    $\mathbf{X} = \{\mathbf{X}_i\} = \{\bm{\rho}_i\}$.
\par \noindent   \textbf{Physical invariant node feature set} $\mathbf{Y}\in \mathbb{R}^{N\times 3}$: This set represents the intrinsic properties of each node $i$ that remain stable under transformation, including the SO(3)-invariant compensated node velocity $v_{i}^{dop^{'}}$, and SE(3)-invariant RCS value $r_i$ and the degree $d_i$, which is the number of edges connected to the node $i$:
    $\mathbf{Y} = \{\mathbf{Y}_i\} =  \{[v_{i}^{dop^{'}}, r_i, d_i]\}$.
\par \noindent \textbf{Relative invariant edge feature set} $\mathbf{Z}\in \mathbb{R}^{E\times 2}$ ($E$ is the total number of edges): To capture local relative information between nodes $i$ and $j$, the edge features consist of the compensated edge relative velocity $v_{ij}$, which remains SO(3)-invariant and further satisfies SE(3)-invariance for nodes with the same velocity, and the SE(3)-invariant Euclidean distance $\left\| \bm{\rho}_i - \bm{\rho}_j\right\|$:
    $\mathbf{Z} = \{\mathbf{Z}_{ij}\} =  \{[v_{ij}, \left\| \bm{\rho}_i - \bm{\rho}_j\right\|]\}, j\in \mathcal{N}(i)$,
where $\mathcal{N}(i)$ is the set of neighbors of $i$ in the graph.

These constructed feature sets ($\mathbf{X}, \mathbf{Y}, \mathbf{Z}$), and the graph structure $\mathbf{G}$, are built independently for both the source and target frames (denoted by subscripts $s$ and $t$ in Fig.~\ref{Fig:overview}) to serve as paired inputs for the subsequent network.

\subsection{Network Design}\label{Subsec:Network}
Following graph construction and feature preparation, we establish a novel framework that processes the input equivariant and invariant features for both the source and target point clouds independently. The framework employs equivariant networks to handle different feature types and progressively integrate geometric structure with physical properties, yielding the respective final per-point feature set $\mathbf{F}_s$ and $\mathbf{F}_t$ for estimating the relative transformation between the two frames.

\subsubsection{Feature Extraction}

We employ a two-stage cascaded architecture for feature extraction of each frame. An SO(3)-equivariant Dynamic Graph Convolutional Neural Network (DGCNN)~\cite{phan2018dgcnn} first processes equivariant node feature set $\mathbf{X}$ of each frame of point clouds and produces SO(3)-equivariant output feature $\mathbf{X'}$. The second module is an SE(3)-equivariant GNN (EGNN) module that, by integrating the processed node feature set $\mathbf{X'}$ with the node and edge invariant feature set $\mathbf{Y}$ and $\mathbf{Z}$, outputs the final feature set $\mathbf{F}$.

The first DGCNN network aggregates features from each point and its neighbors by a local graph, generating richer local geometric representations. Before applying the network, we compensate the dominant translational component using ego-velocity and time interval, so that the SO(3)-equivariant layers can focus on rotation-consistent feature learning, and the whole system predicts the residual SE(3) motion on top of the translational initialization. A stack of LN~\cite{lin2024lie} network layers is applied in the DGCNN framework. LN modifies traditional MLP layers such as Linear, ReLU and Pooling layers for equivariance on semi-simple Lie groups, and preserves SO(3)-equivariance in our case. After three iterations, the network outputs the processed node geometric feature set $\mathbf{X'}$, whose dimension is lifted to $N\times(C'\times3)$ by LN and retains SO(3)-equivariance with SE(3)-aware pose reasoning.

The processed node feature set $\mathbf{X'}$ of each frame is then passed to the second EGNN module together with physical invariant node feature set $\mathbf{Y}$ and relative invariant edge feature set $\mathbf{Z}$ of this frame for further graph feature aggregation. For each node $i$ and its neighbors $\mathcal{N}(i)$ connected by edges, the features are aggregated and updated as follows:
\begin{align}
    \mathbf{m}_{ij} &= \phi_{I_{1}} (\exp(\gamma\left\|\mathbf{X}^{'}_{i}-\mathbf{X}^{'}_{j}\right\|),\mathbf{Y}_{i},\mathbf{Y}_{j}, \mathbf{Z}_{ij}), \\
    \mathbf{Y}_{i} &= \phi_{I_{2}} (\mathbf{Y}_{i}, \underset{j\in \mathcal{N}(i)}{\sum} \mathbf{m}_{ij}), \\
    \mathbf{X}^{'}_{i} &= \mathbf{X}^{'}_{i} + \frac{1}{\left|\mathcal{N}(i)\right|} \underset{j\in \mathcal{N}(i)}{\sum} ( \mathbf{X}^{'}_{i}- \mathbf{X}^{'}_{j}) \phi_{E}(\mathbf{m}_{ij}),
\end{align}
where $\mathbf{m}_{ij}$ represents the invariant message passed from node $j$ to $i$ under the message passing neural network framework~\cite{gilmer2017neural}, $\mathbf{X}^{'}_{i}$ and $\mathbf{Y}_{i}$ are processed equivariant and invariant features of node $i$, respectively. $\phi_{I_{1}}, \phi_{I_{2}}, \phi_{E}$ are MLPs. 

Differing from the original EGNN that uses $\left\|\mathbf{X}^{'}_{i} - \mathbf{X}^{'}_{j}\right\|^2$ for invariant feature extraction, inspired by~\cite{zhang2024correspondence}, we employ a Gaussian kernel function $\exp(\gamma \left\|\mathbf{X}_{i}^{'} - \mathbf{X}_{j}^{'}\right\|)$ with a learnable parameter $\gamma$ to transform the SO(3)-equivariant features $\mathbf{X'}$ into SE(3)-invariant representations.

After four iterations of the EGNN, the network outputs final node equivariant feature set $\mathbf{X}_{F} = \{\mathbf{X}^{'}_{i}\}$ and node invariant feature set $\mathbf{Y}_{F} = \{\mathbf{Y}_{i}\}$ for $i=1,2,...,N$ for the processed frame. Each frame's final per-point feature set $\mathbf{F}$ is defined as the concatenation of $\mathbf{X}_{F}$ and $\mathbf{Y}_{F}$.

\subsubsection{Transformation Estimation}
In this module, the $\mathbf{L}_2$ distances are computed between every feature vector $\mathbf{f}_{s,i}$ and $\mathbf{f}_{t,j}$ in the source and target feature sets $\mathbf{F}_s$ and $\mathbf{F}_t$, respectively, forming a pairwise-correspondence matrix. The $M$ pairs with the smallest feature distances are selected as keypoint correspondences. A learnable similarity metric is then defined as:
    $s_{ij} = \exp(-\beta(d_{ij}-\alpha))$,
where $s_{ij}$ denotes the similarity between source point $i$ and target point $j$, $d_{ij}$ is their feature distance, and $\alpha, \beta$ are learnable parameters that enhance the robustness to outliers.

Given the similarity matrix $\mathbf{S} = \{s_{ij}\}_{M \times M}$, we obtain a soft assigning weight matrix $\mathbf{W}=\{w_{ij}\}_{M\times M}$ of correspondences via the differentiable Sinkhorn algorithm~\cite{sinkhorn1967concerning}. A weighted Singular Value Decomposition (SVD) finally estimates the transformation $(\bm{R}, \bm{t})$ between the two frames.

\subsubsection{Loss Function}
The proposed framework is trained in an end-to-end manner using the ground truth transformation between frames as supervision. Specifically, we design a novel loss function that jointly considers rotation, translation, and orientation errors. The estimated rotation $\bm{R}$ is first converted to Euler angles $\mathbf{E}$, and the loss $\mathcal{L}$ between $(\mathbf{E}, \mathbf{t})$ and the ground truth $(\mathbf{E}_{gt}, \mathbf{t}_{gt})$ consists of four terms: rotation loss $\mathcal{L}_r$, translation loss $\mathcal{L}_t$, pitch loss $\mathcal{L}_p$, and yaw loss $\mathcal{L}_y$:
\begin{align}
    \nonumber \mathcal{L}_r &= \left\| \mathbf{E}-\mathbf{E}_{gt}\right\|, \mathcal{L}_t = \left\| \mathbf{t}-\mathbf{t}_{gt}\right\|, \\
    \mathcal{L}_p&= \left\| \mathbf{E}^{pitch}-\mathbf{E}_{gt}^{pitch}\right\|, 
    \mathcal{L}_y = \left\| \mathbf{E}^{yaw}-\mathbf{E}_{gt}^{yaw}\right\|,
\end{align}
where $\mathbf{E}^{pitch}$ and $\mathbf{E}_{gt}^{pitch}$ denote the pitch angles, and $\mathbf{E}^{yaw}$ and $\mathbf{E}_{gt}^{yaw}$ denote the yaw angles, in the estimated and ground truth rotations, respectively. We include additional pitch loss $\mathcal{L}_p$ and yaw loss $\mathcal{L}_y$ for finer-grained rotation supervision. Geometrically, pitch and yaw directly affect forward motion, whereas roll primarily influences lateral and vertical components, which have less impact on forward trajectory error and can be sufficiently constrained by the main rotation loss $\mathcal{L}_r$.

Following~\cite{zhuo20234drvo}, we introduce four learnable balancing parameters $s_r, s_t, s_p, s_y$ to account for the different units and scales of rotation and translation, leading to the final loss:
\begin{align}
   \notag  \mathcal{L} &= \mathcal{L}_r \exp(-s_r) + s_r + \mathcal{L}_t \exp(-s_t) + s_t \\
    &+ \mathcal{L}_p \exp(-s_p) + s_p + \mathcal{L}_y \exp(-s_y) + s_y.
\end{align}

\section{Experiments}  \label{Sec:Experiments}
In this section, we present comprehensive experiments to evaluate our proposed Equi-RO framework.
\subsection{Datasets}
Two datasets are used for evaluation: the NTU4DRadLM dataset~\cite{zhang2023ntu4dradlm} and a self-collected dataset acquired using a vehicle-mounted 4D radar and high-precision GPS.

\subsubsection{NTU4DRadLM Dataset}
This dataset is designed for 4D radar-based localization and mapping. It provides ground-truth odometry and includes sequences from both low-speed robot platforms and high-speed vehicles. It was collected using the Oculii Eagle 4D radar, which has $0.5^\circ$ azimuth and $1^\circ$ elevation resolution at 12 Hz operating frequency. We use its enhanced radar point cloud and select \textit{loop1} for training, \textit{loop2} for validation, and the remaining four splits for testing.

\subsubsection{Self-Collected Dataset}
\begin{figure}[t]
    \centering
    \includegraphics[width=0.23\textwidth]{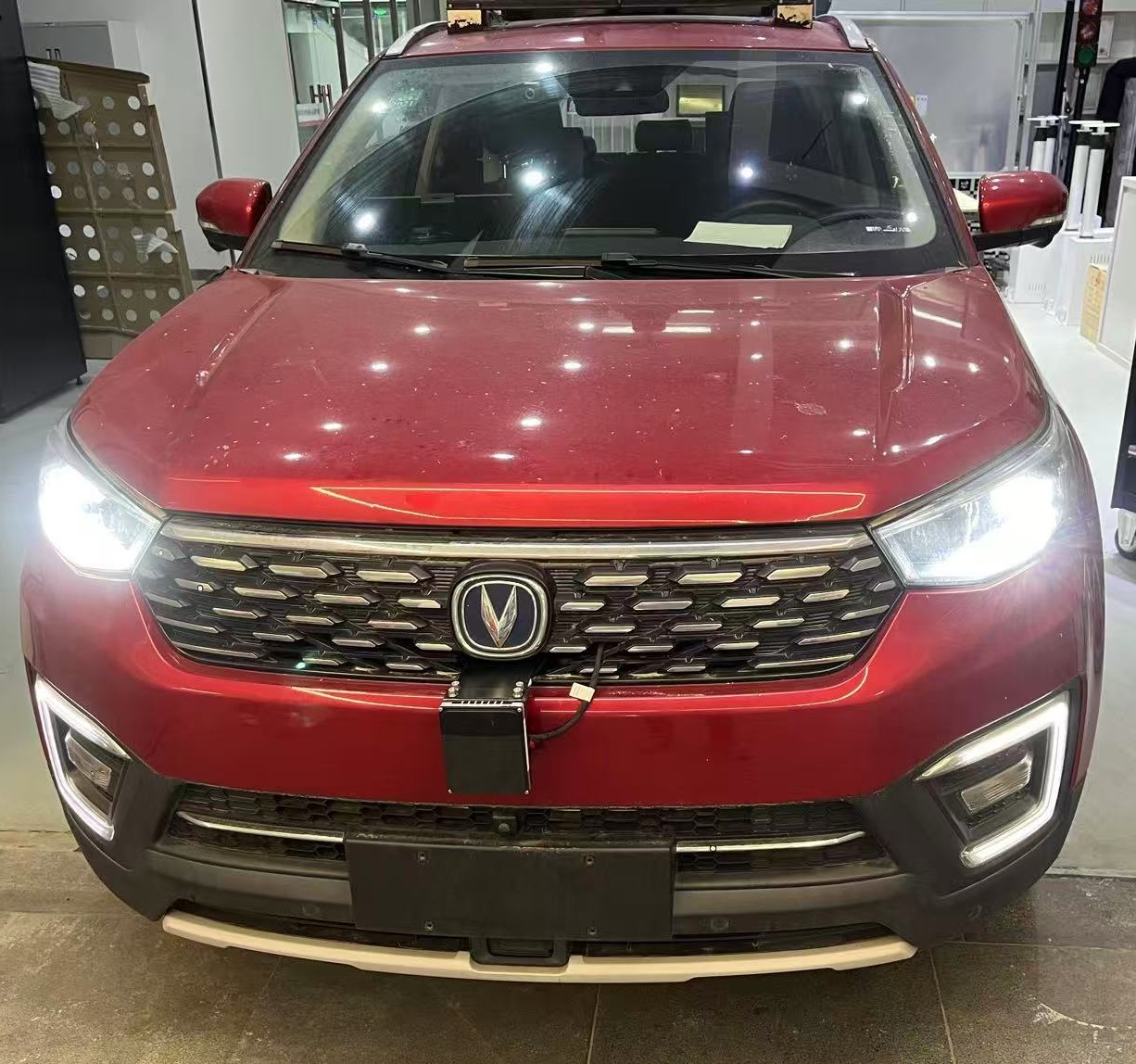}
    \caption{Test vehicle equipped with a 4D radar for the self-collected dataset.}
    \label{Fig:self-vehicle}
    \vspace{-5mm}
\end{figure}

To further assess generalization, we constructed a campus dataset using the Geometrical Pal R7861B 4D radar. As shown in Fig.~\ref{Fig:self-vehicle}, the radar was mounted on an SUV equipped with a high-precision GPS system to provide ground-truth odometry. The radar has $1.9^\circ$ azimuth and $3.5^\circ$ elevation resolution, operates at 15 Hz. All learning-based models trained on NTU4DRadLM are directly tested on this dataset to compare cross-dataset generalization.

\begin{figure}[tb]
    \centering
    \includegraphics[width=0.48\textwidth]{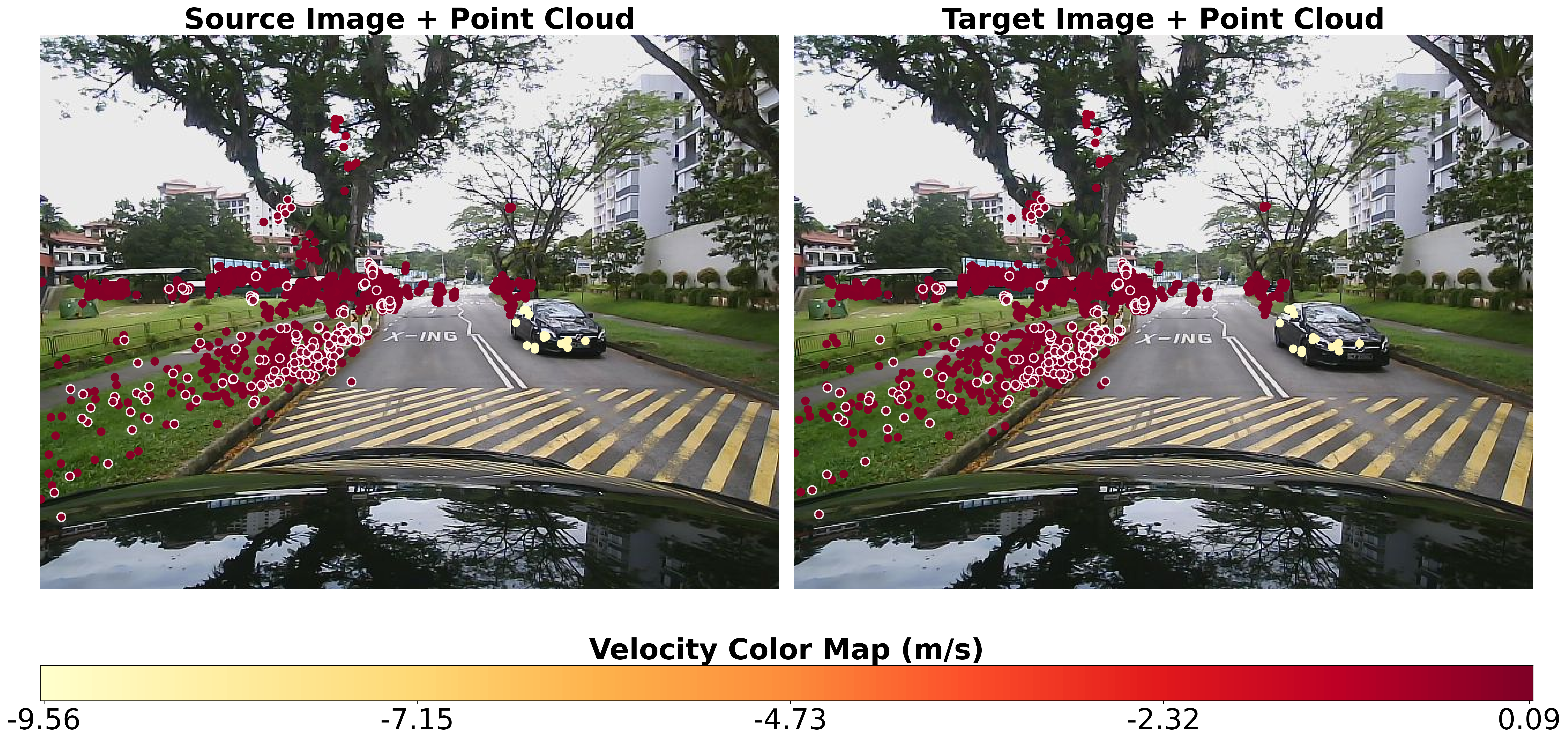}
    \caption{Projection of adjacent point clouds onto the corresponding images from \textit{loop3} split, where point colors represent compensated node velocities and white circles denote selected keypoints used for matching.}
    \label{Fig:quali_vis}
    \vspace{-5mm}
\end{figure}

\begin{figure*}[!t]
    \centering
    \subfigure[]{
    \includegraphics[width=0.23\textwidth]{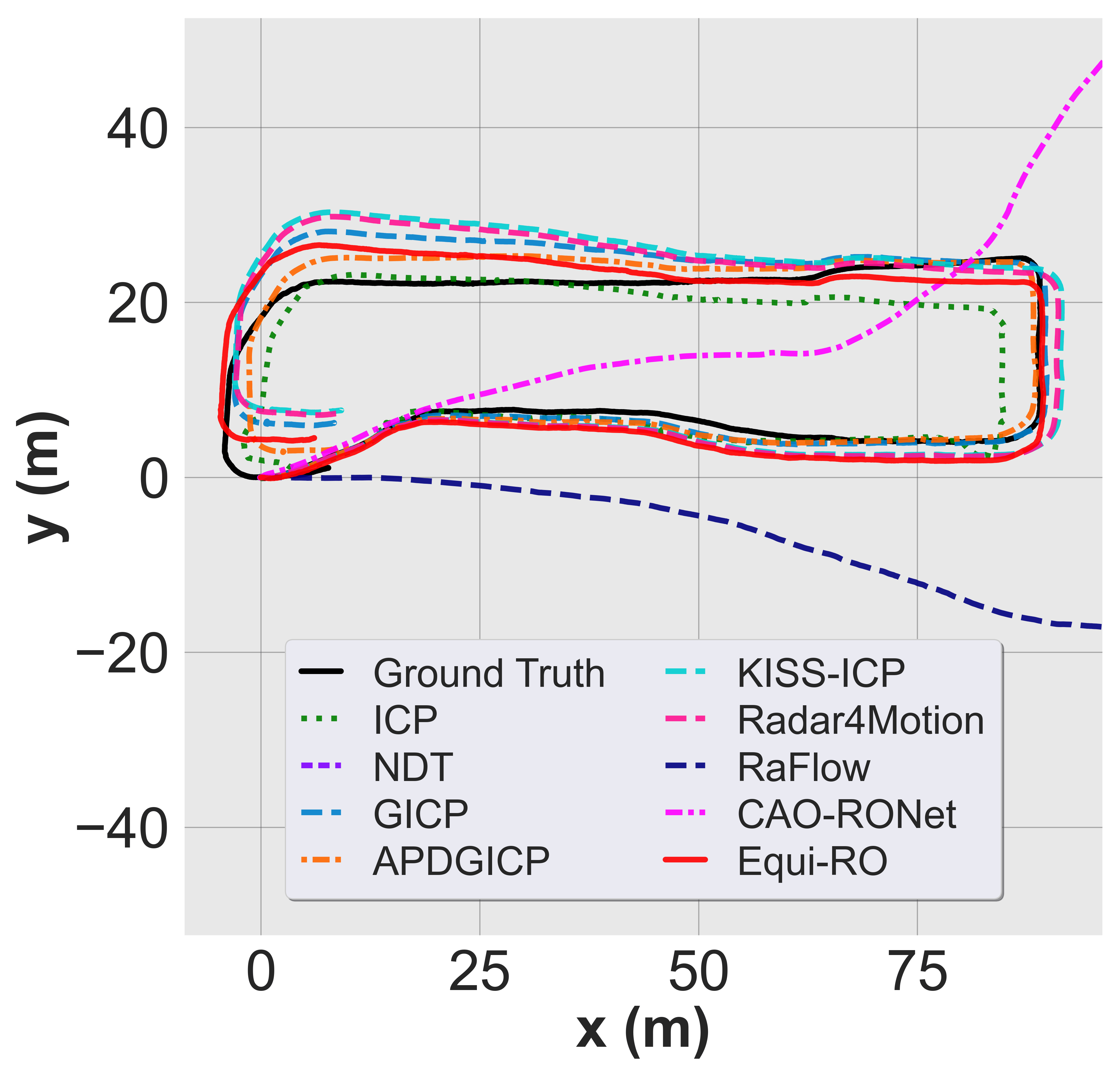}
    \label{Fig:cp}}\subfigure[]{
    \includegraphics[width=0.23\textwidth]{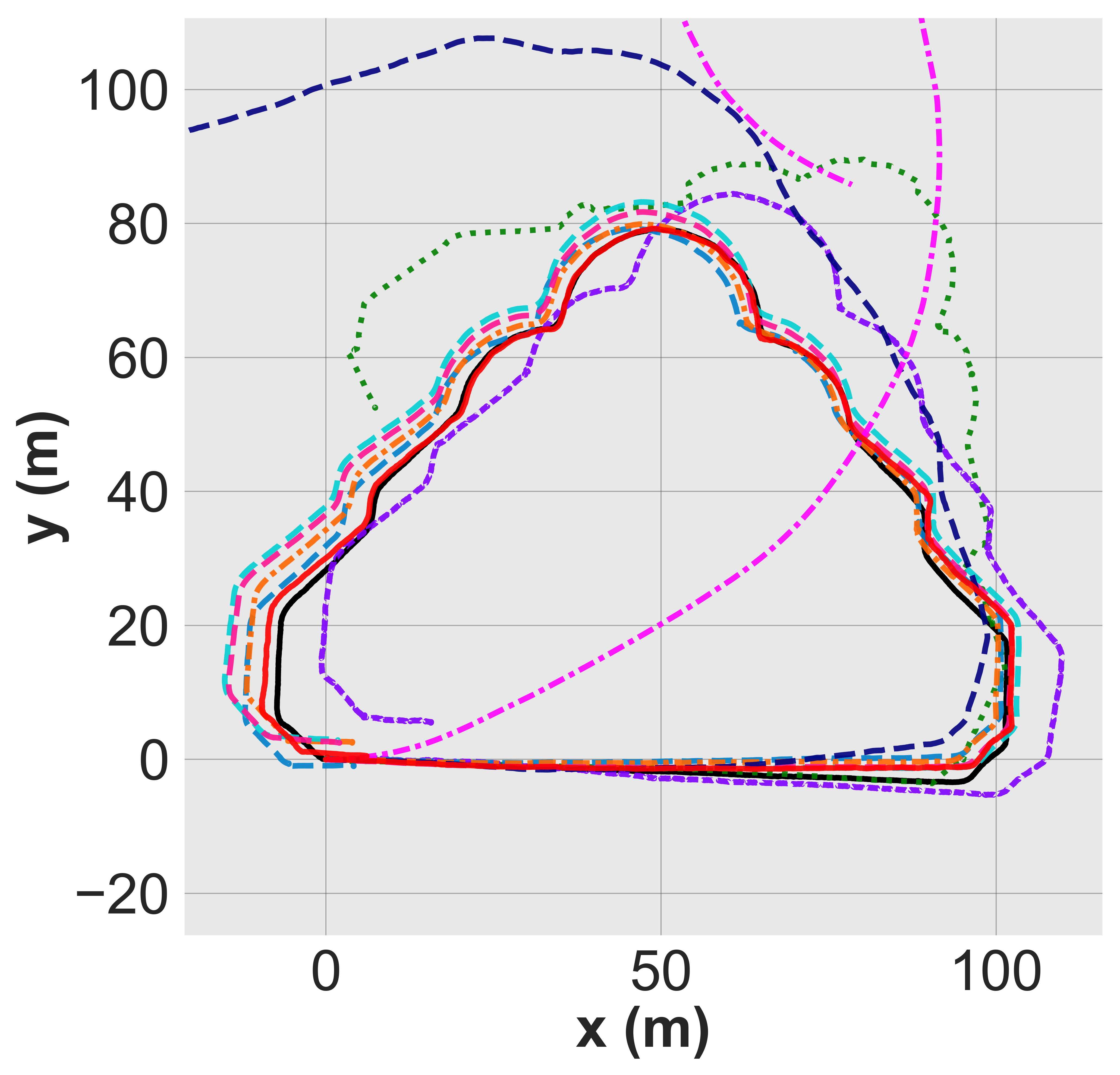}
    \label{Fig:garden}}\subfigure[]{
    \includegraphics[width=0.24\textwidth]{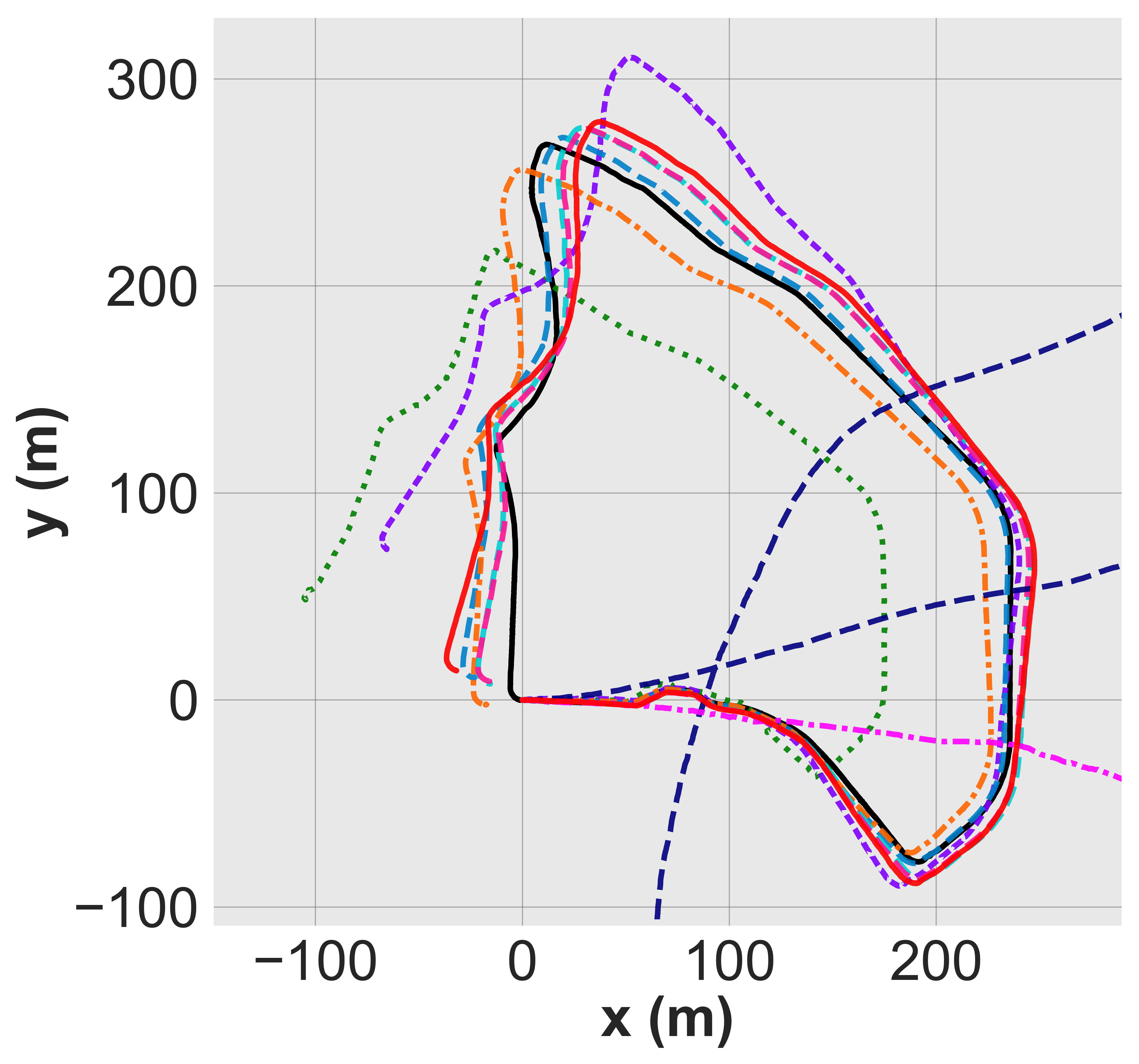}
    \label{Fig:nyl}}\subfigure[]{
    \includegraphics[width=0.24\textwidth]{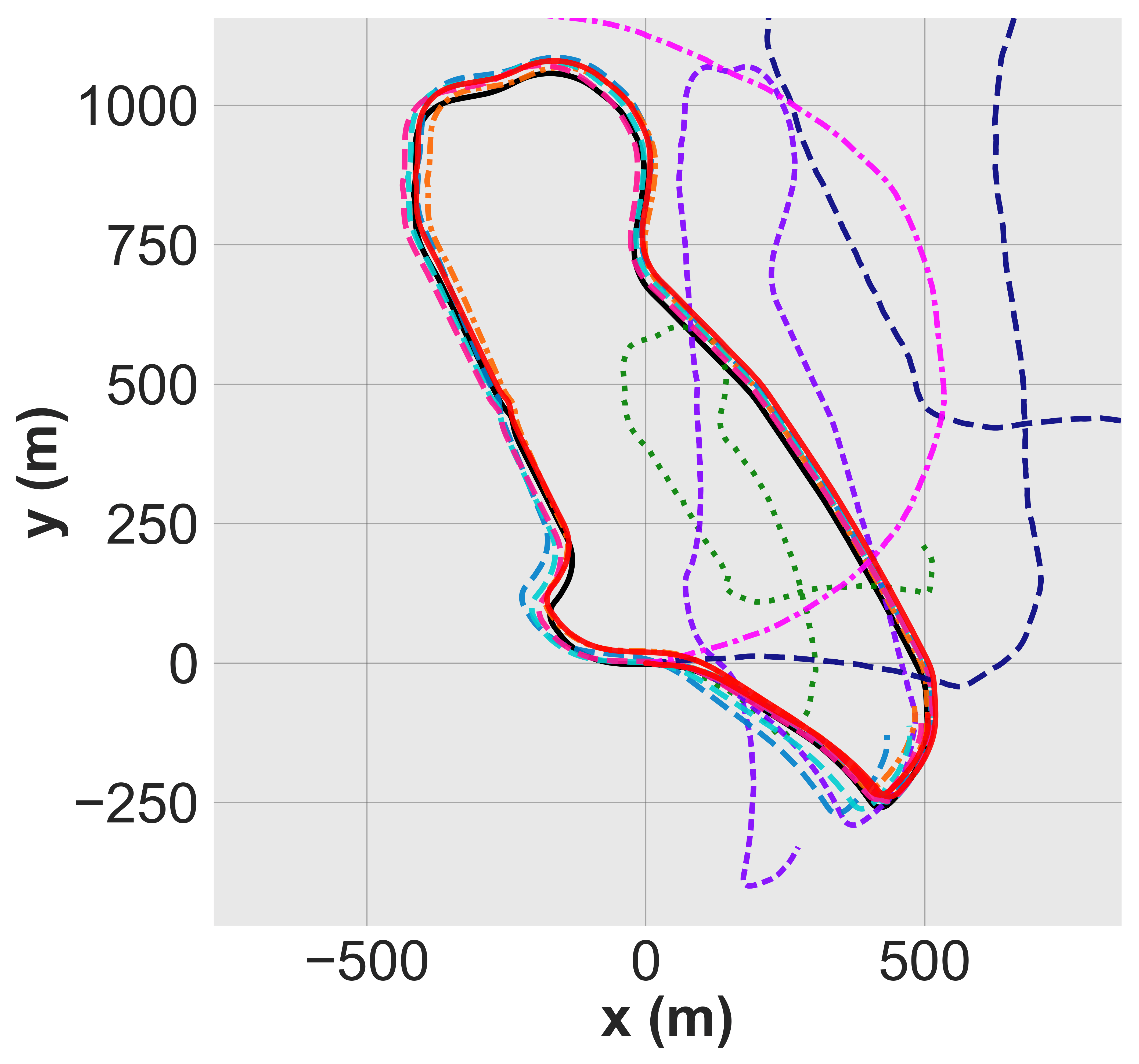}
    \label{Fig:loop3}}
    \caption{Comparative odometry results of Equi-RO and baselines on the NTU4DRadLM dataset for the \textit{cp} (a), \textit{garden} (b), \textit{nyl} (c), and \textit{loop3} (d) splits.}
    \label{Fig:vis}
\vspace{-3mm}
\end{figure*}

\begin{table*}[!t]
\belowrulesep=0pt
\aboverulesep=0pt
\centering
\caption{Quantitative comparison on NTU4DRadLM and Self-collected datasets ($t_{\text{rel}}:\%, r_{\text{rel}}:^\circ/\text{m}$)}
\label{Tab:results}
\resizebox{\linewidth}{!}{
\begin{tabular}{@{}ccccccccccc!{\vrule}ccccccc@{}}
\toprule
 & \multicolumn{10}{c!{\vrule}}{\textbf{NTU4DRadLM}} & \multicolumn{7}{c}{\textbf{Self-collected}} \\
\cmidrule(lr){2-11} \cmidrule(lr){12-18}
Method & \multicolumn{2}{c}{cp} & \multicolumn{2}{c}{garden} & \multicolumn{2}{c}{nyl} & \multicolumn{2}{c}{loop3} & \multicolumn{2}{c!{\vrule}}{Average} 
       & \multicolumn{2}{c}{1} & \multicolumn{2}{c}{2} & \multicolumn{2}{c}{Average} \\
\cmidrule(lr){2-3} \cmidrule(lr){4-5} \cmidrule(lr){6-7} \cmidrule(lr){8-9} \cmidrule(lr){10-11} \cmidrule(lr){12-13} \cmidrule(lr){14-15} \cmidrule(lr){16-17}
 & \(t_{\text{rel}}\) & \(r_{\text{rel}}\) 
 & \(t_{\text{rel}}\) & \(r_{\text{rel}}\) 
 & \(t_{\text{rel}}\) & \(r_{\text{rel}}\) 
 & \(t_{\text{rel}}\) & \(r_{\text{rel}}\) 
 & \(t_{\text{rel}}\) & \(r_{\text{rel}}\) 
 & \(t_{\text{rel}}\) & \(r_{\text{rel}}\) 
 & \(t_{\text{rel}}\) & \(r_{\text{rel}}\) 
 & \(t_{\text{rel}}\) & \(r_{\text{rel}}\) \\
\midrule
ICP        & 6.88 & 0.0774 & 17.24 & 0.1997 & 20.05 & 0.0635 & 50.43 & 0.0461 & 23.15 & 0.0967 
           & 54.15 & 0.0698 & 63.89 & 0.3721 & 59.02 & 0.2209 \\
NDT        & Failed & Failed & 5.46 & 0.0818 & 8.86 & 0.0464 & 13.10 & 0.0255 & 9.14 & 0.0512 
           & Failed & Failed & Failed & Failed & Failed & Failed \\
GICP       & \underline{4.09} & 0.0576 & \underline{3.27} & 0.0459 & 4.39 & 0.0248 & 4.99 & 0.0081 & 4.19 & 0.0341 
           & 41.23 & 0.0621 & 49.89 & \underline{0.0814} & 45.56 & \underline{0.0718} \\
APDGICP    & 4.31 & 0.0575 & 3.33 & 0.0472 & \textbf{3.95} & \textbf{0.0210} & 4.81 & 0.0100 & \underline{4.10} & 0.0340 
           & \underline{10.97} & 0.0575 & 34.83 & 0.0961 & 22.90 & 0.0768 \\
KISS-ICP   & 4.86  & \underline{0.0486} & 4.38  & 0.0439 & 4.34  & 0.0237 & \underline{3.88} & 0.0092 & 4.37 & \underline{0.0314} & 11.07 & \underline{0.0563} & 25.07 & 0.0879 & \underline{18.07} & 0.0721 \\
Radar4Motion &4.65 & 0.0565 & 3.87  & \underline{0.0395} & \underline{4.11}  & 0.0265 & 4.14 & \underline{0.0080} & 4.19 & 0.0326 & 13.62 & 0.0605 & \underline{22.59} & 0.0858 & 18.11 & 0.0732 \\
RaFlow     & 62.12 & 0.2362 & 37.07 & 0.3961 & 25.93 & 0.1060 & 60.31 & 0.1765 & 46.86 & 0.2287 
           & 58.43 & 0.0846 & 67.59 & 0.1296 & 63.01 & 0.1071 \\
CAO-RONet  & 73.31 & 0.9404 & 50.42 & 0.5898 & 29.81 & 0.2179 & 78.36 & 0.1978 & 57.98 & 0.4865 
           & 73.31 & 0.0994 & 55.54 & 0.2781 & 64.42 & 0.1888 \\
\textbf{Equi-RO} 
           & \textbf{3.93} & \textbf{0.0421} & \textbf{3.15} & \textbf{0.0380} & 4.31 & \underline{0.0218} & \textbf{3.25} & \textbf{0.0070} & \textbf{3.66} & \textbf{0.0272} 
           & \textbf{8.55} & \textbf{0.0524} & \textbf{10.23} & \textbf{0.0711} & \textbf{9.39} & \textbf{0.0617} \\
\bottomrule
\end{tabular}}
\vspace{-5mm}
\end{table*}

\begin{figure}[!t]
    \centering
    \subfigure[]{
        \includegraphics[width=0.215\textwidth]{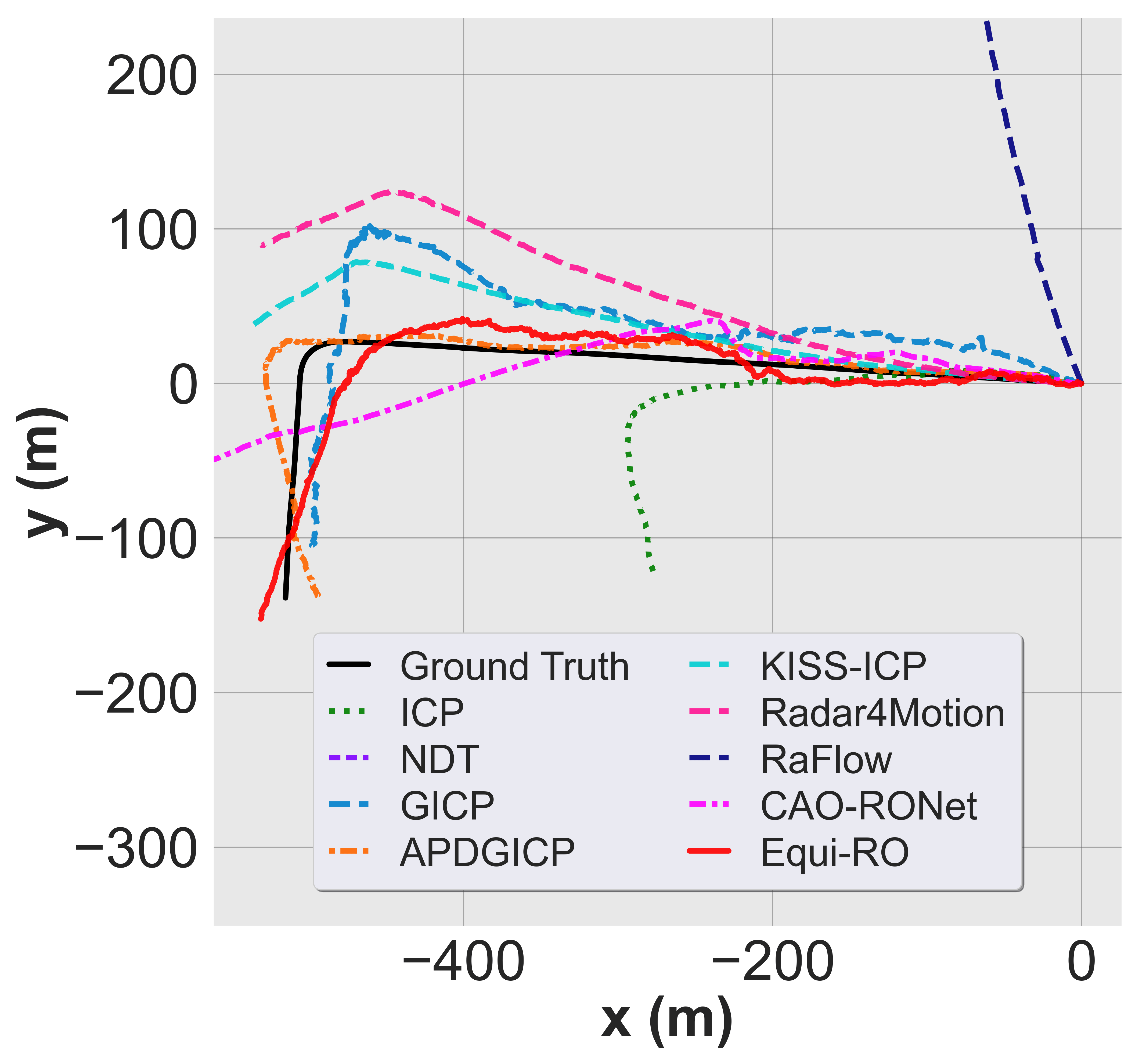}
        \label{Fig:Geo1}}\subfigure[]{
        \includegraphics[width=0.218\textwidth]{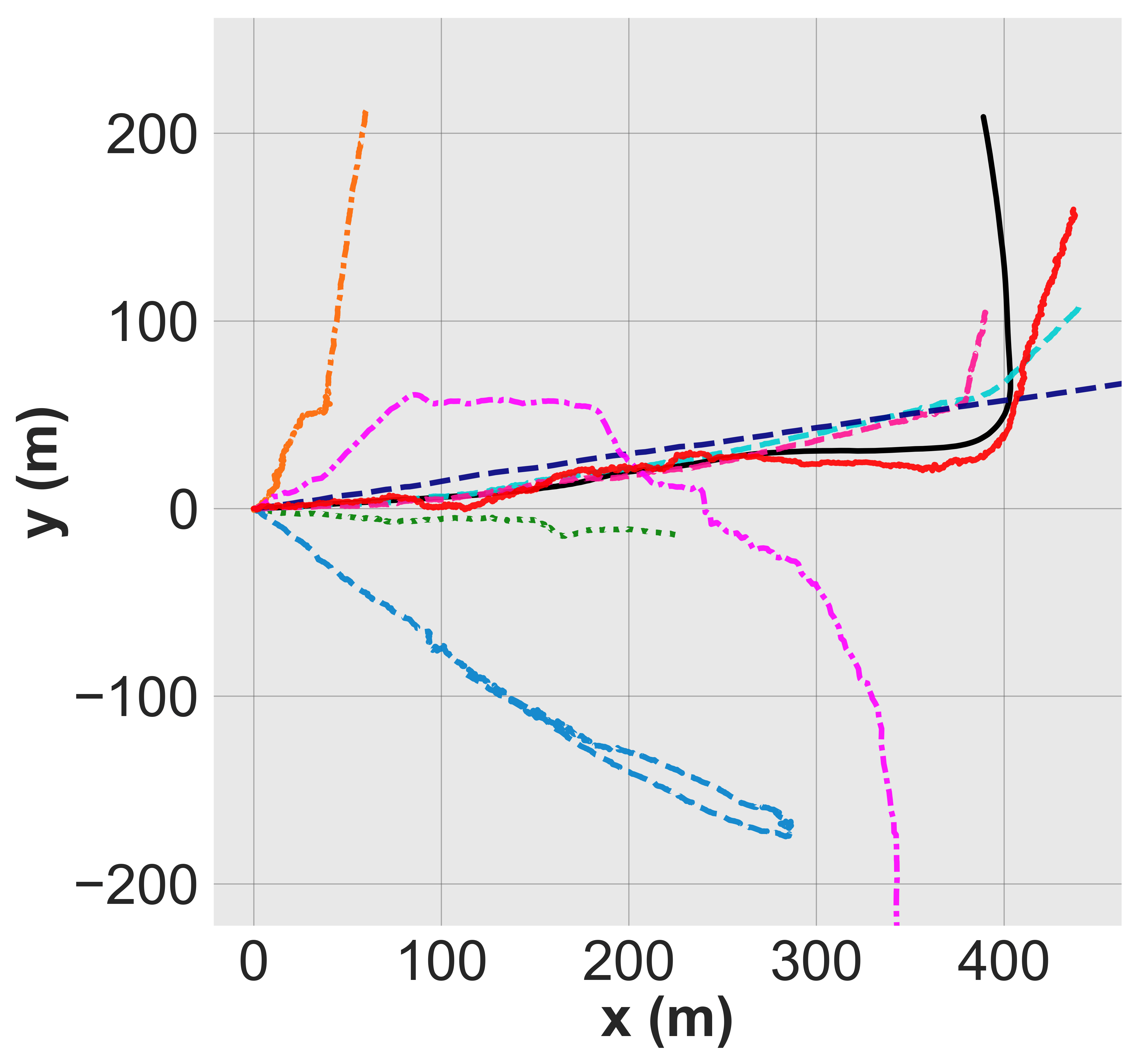}
        \label{Fig:Geo2}}
    \caption{Comparative odometry results of Equi-RO and baselines on the two splits of self-collected dataset.}
    \label{Fig:self-vis}
\vspace{-3mm}
\end{figure}

\begin{figure}[tb]
    \centering
    \subfigure[]{
    \includegraphics[width=0.22\textwidth]{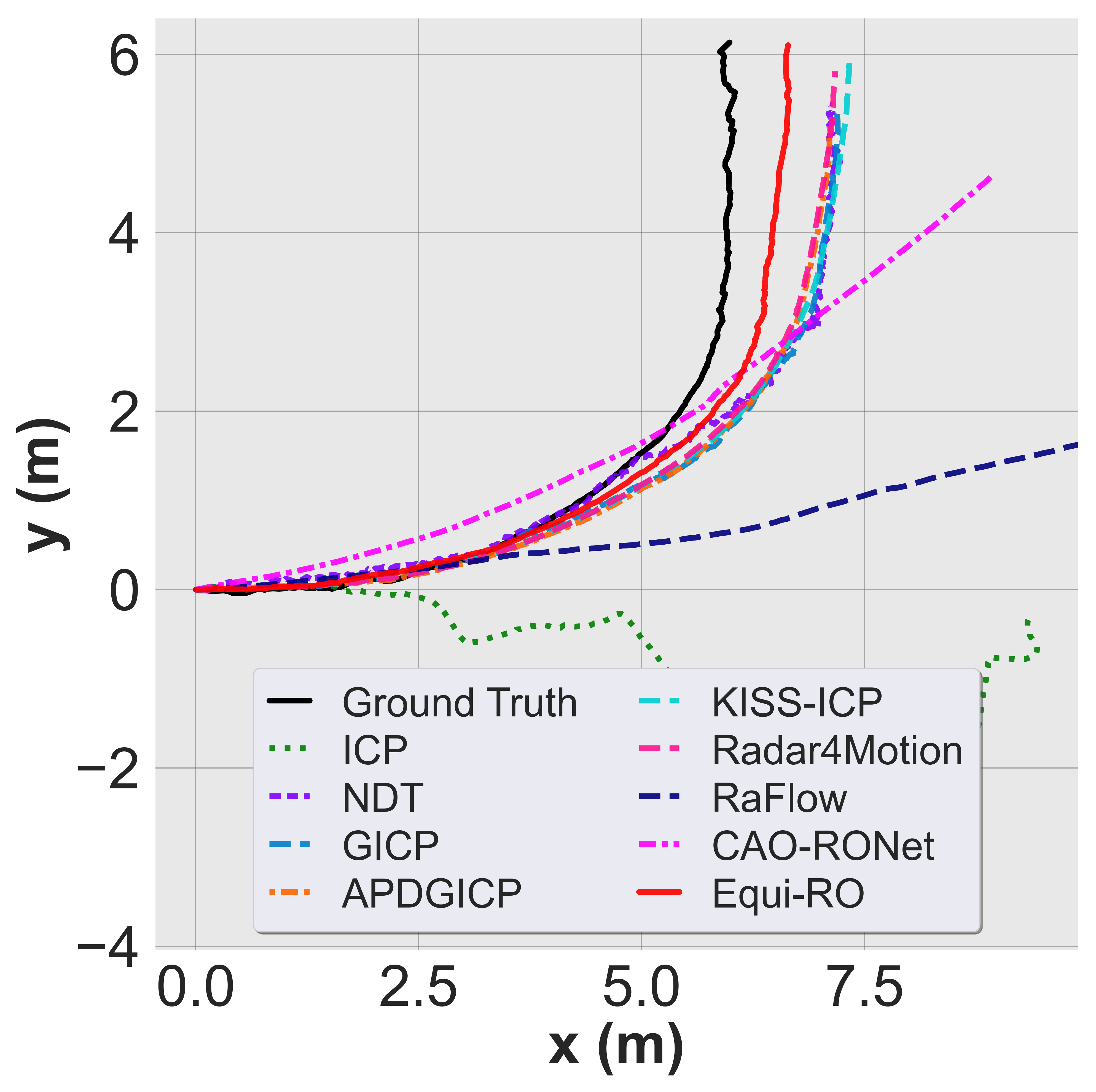}
    \label{Fig:cp_seg}}\subfigure[]{
    \includegraphics[width=0.22\textwidth]{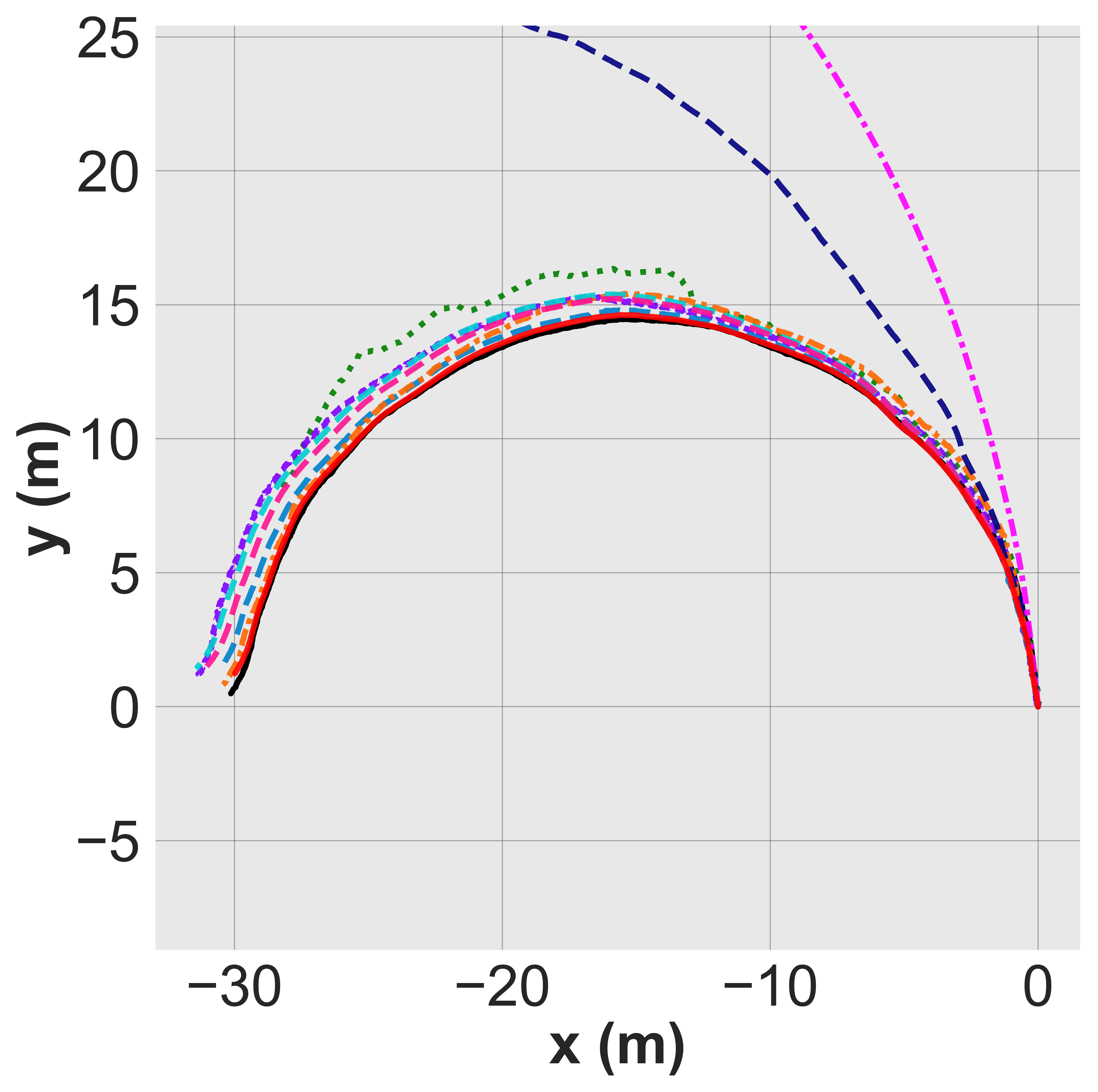}
    \label{Fig:garden_seg}}
    \subfigure[]{
    \includegraphics[width=0.22\textwidth]{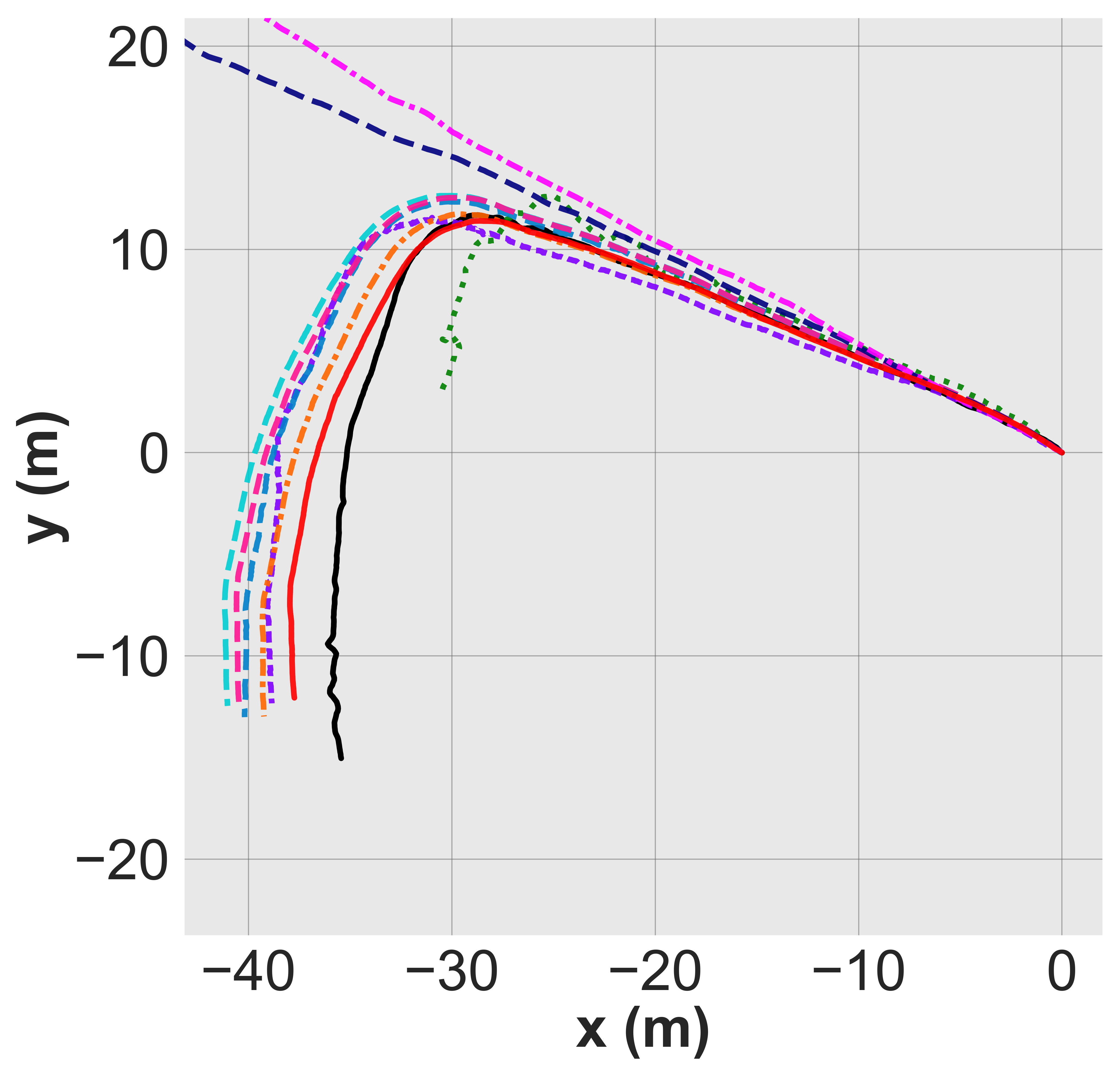}
    \label{Fig:nyl_seg}}\subfigure[]{
    \includegraphics[width=0.22\textwidth]{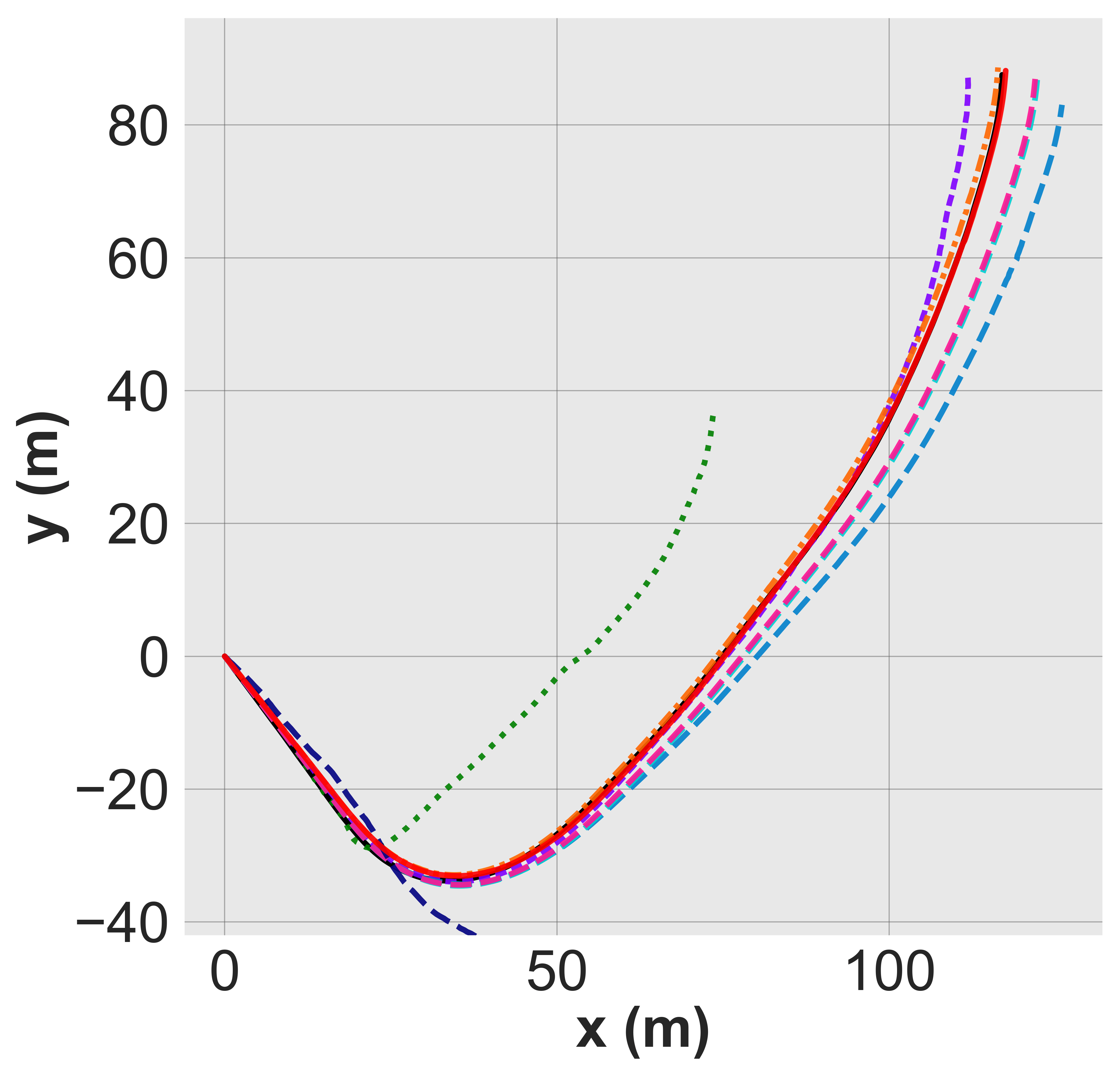}
    \label{Fig:loop3_seg}}
    \caption{Comparative odometry results on large-rotation segments of \textit{cp} (a), \textit{garden} (b), \textit{nyl} (c), and \textit{loop3} (d) splits of the NTU4DRadLM dataset.}
    \label{Fig:largeR}
\vspace{-5mm}
\end{figure}

\begin{table}[tb]
    \centering
    \caption{Quantitative results on large-rotation segments of the NTU4DRadLM dataset}
    \label{Tab:largeR}
    \resizebox{\linewidth}{!}{
    \begin{tabular}{@{}ccccccccc@{}}
    \toprule
    \multirow{2}{*}{Method} & \multicolumn{2}{c}{cp} & \multicolumn{2}{c}{garden} & \multicolumn{2}{c}{nyl} & \multicolumn{2}{c}{loop3} \\
     \cmidrule(lr){2-3} \cmidrule(lr){4-5} \cmidrule(lr){6-7} \cmidrule(lr){8-9}
     & \( t_{\text{rel}} \) & \( r_{\text{rel}} \) & \( t_{\text{rel}} \) & \( r_{\text{rel}} \) & \( t_{\text{rel}} \) & \( r_{\text{rel}} \) & \( t_{\text{rel}} \) & \( r_{\text{rel}} \) \\
    \midrule
    ICP          & 70.04    & 0.6056    & 20.59   & 0.2884    & 63.04   & 0.2001 & 53.81   & 0.5660        \\
    NDT          & Failed   & Failed    & 5.76    & \textbf{0.1337}    & 7.23    & 0.0881 & 6.84    & 0.5276        \\
    GICP         & 16.51    & 0.5375    & 4.73    & 0.1706    & 8.44    & 0.0816 & 6.68    & 0.5043        \\
    APDGICP      & 17.61    & 0.5189    & \underline{3.63}    & 0.1578    & \underline{6.83}    & 0.0846 & \underline{4.94}    & \underline{0.4480}        \\
    KISS-ICP     & 14.35    & \underline{0.4219}    & 5.33    & 0.1785    & 7.23    & 0.0982 & 7.23    & 0.5150        \\
    Radar4Motion & \underline{13.26}    & 0.4398    & 4.86    & 0.1813    & 7.38    & \textbf{0.0738} & 6.52    & 0.5109        \\
    RaFlow       & 50.43    & 2.1401    & 43.78   & 0.5364    & 37.14   & 0.2808 & 60.96   & 0.9989        \\
    CAO-RONet    & 29.30    & 1.9730    & 50.45   & 0.2681    & 28.83   & 0.3657 & 72.31   & 1.6608        \\
    Equi-RO      & \textbf{7.23}     & \textbf{0.2588}    & \textbf{3.38}    & \underline{0.1457}    & \textbf{6.47}    & \underline{0.0762} & \textbf{4.64}    & \textbf{0.4405}        \\
    \bottomrule
    \end{tabular}}
    
\vspace{-5mm}
\end{table}

\begin{figure*}[!t]
    \centering
    \subfigure[]{
        \includegraphics[width=0.22\textwidth]{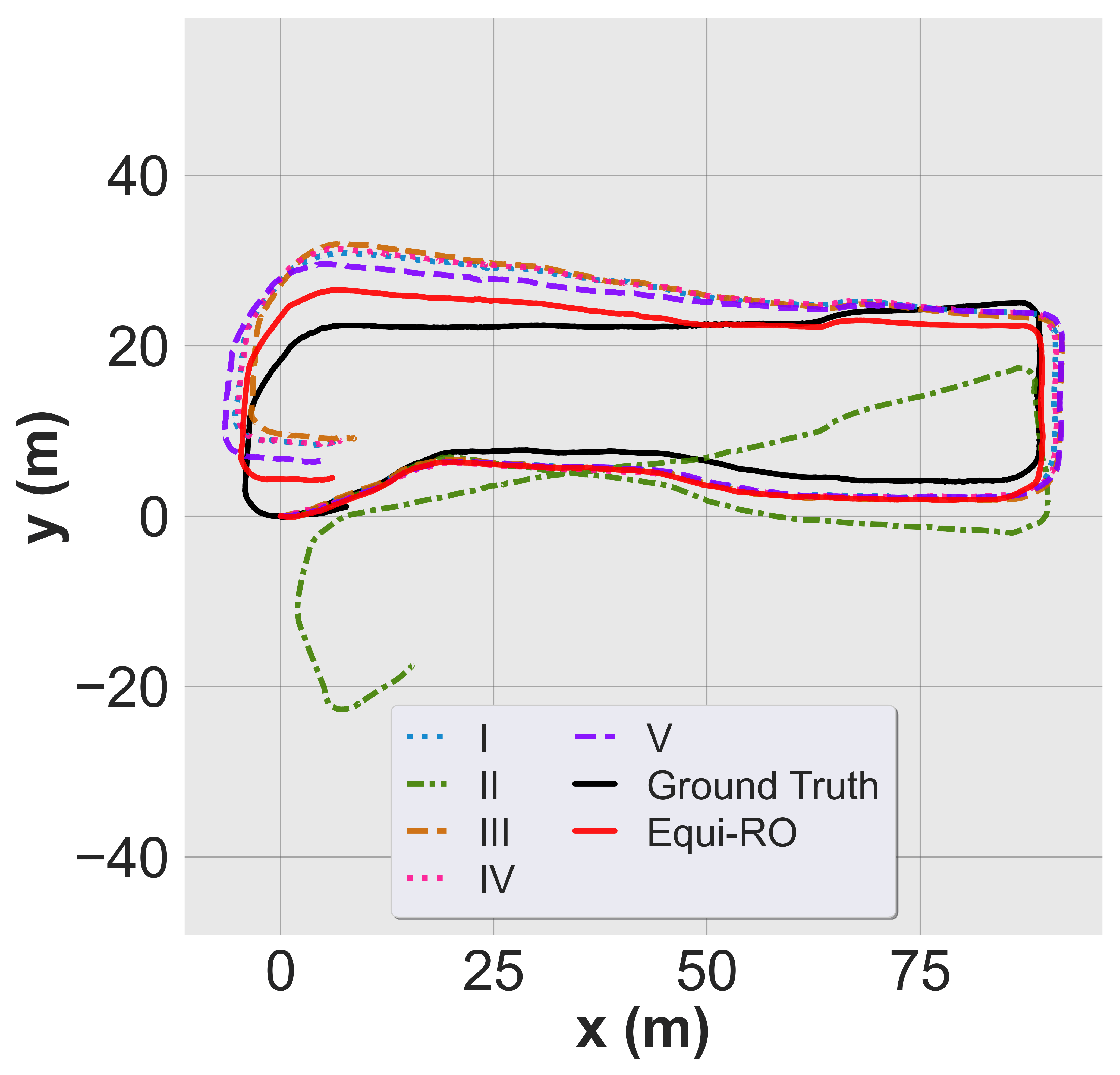}
        \label{Fig:abla_cp}}\subfigure[]{
        \includegraphics[width=0.22\textwidth]{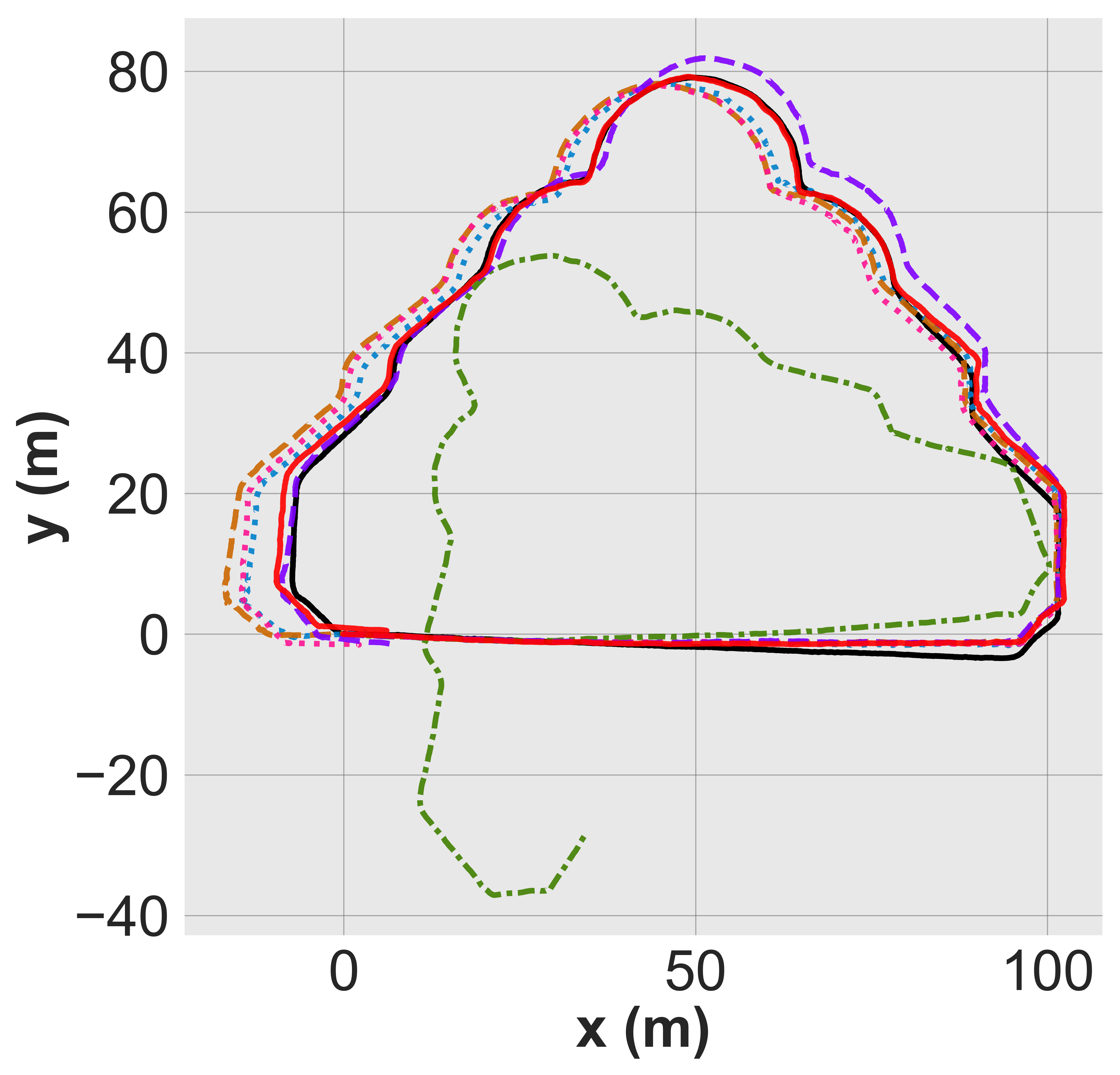}
        \label{Fig:abla_garden}}\subfigure[]{
        \includegraphics[width=0.23\textwidth]{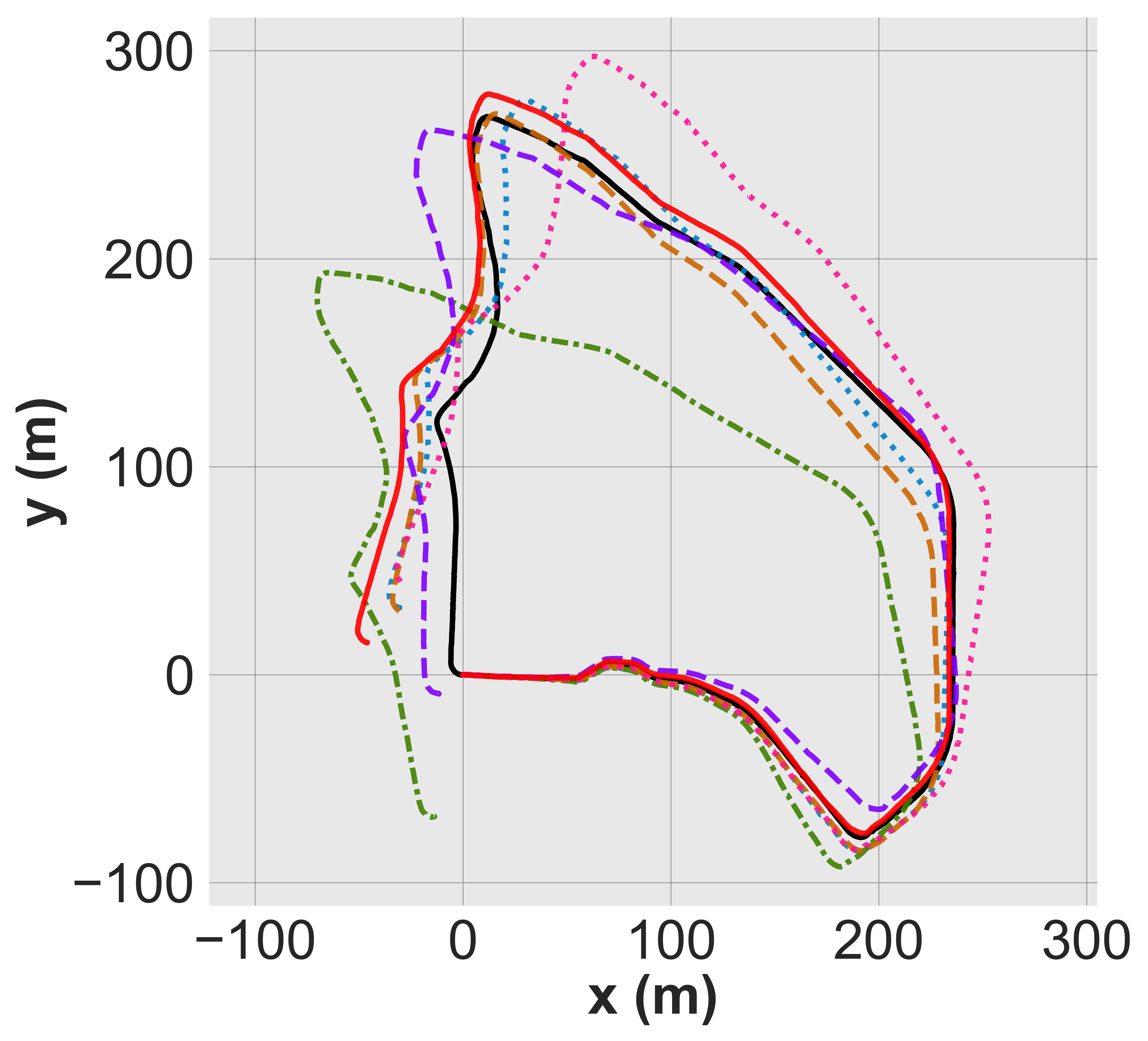}
        \label{Fig:abla_nyl}}\subfigure[]{
        \includegraphics[width=0.22\textwidth]{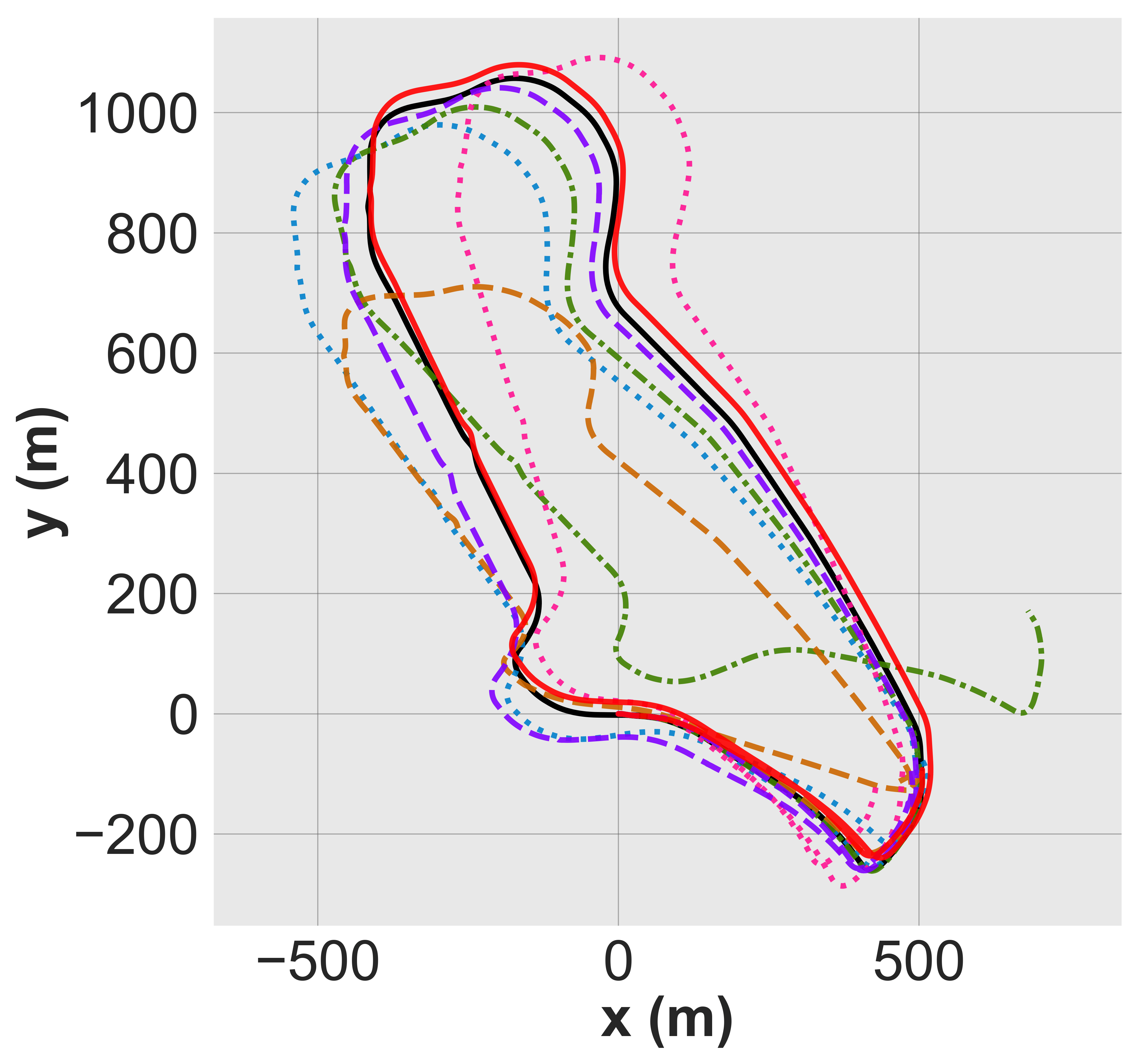}
        \label{Fig:abla_loop3}}
    \caption{Ablation study results of Equi-RO on the \textit{cp} (a), \textit{garden} (b), \textit{nyl} (c), and \textit{loop3} (d) splits of the NTU4DRadLM dataset.}
    \label{Fig:ablation}
\vspace{-5mm}
\end{figure*}

\begin{figure}[tb]
    \centering
    \subfigure{
        \includegraphics[width=0.23\textwidth]{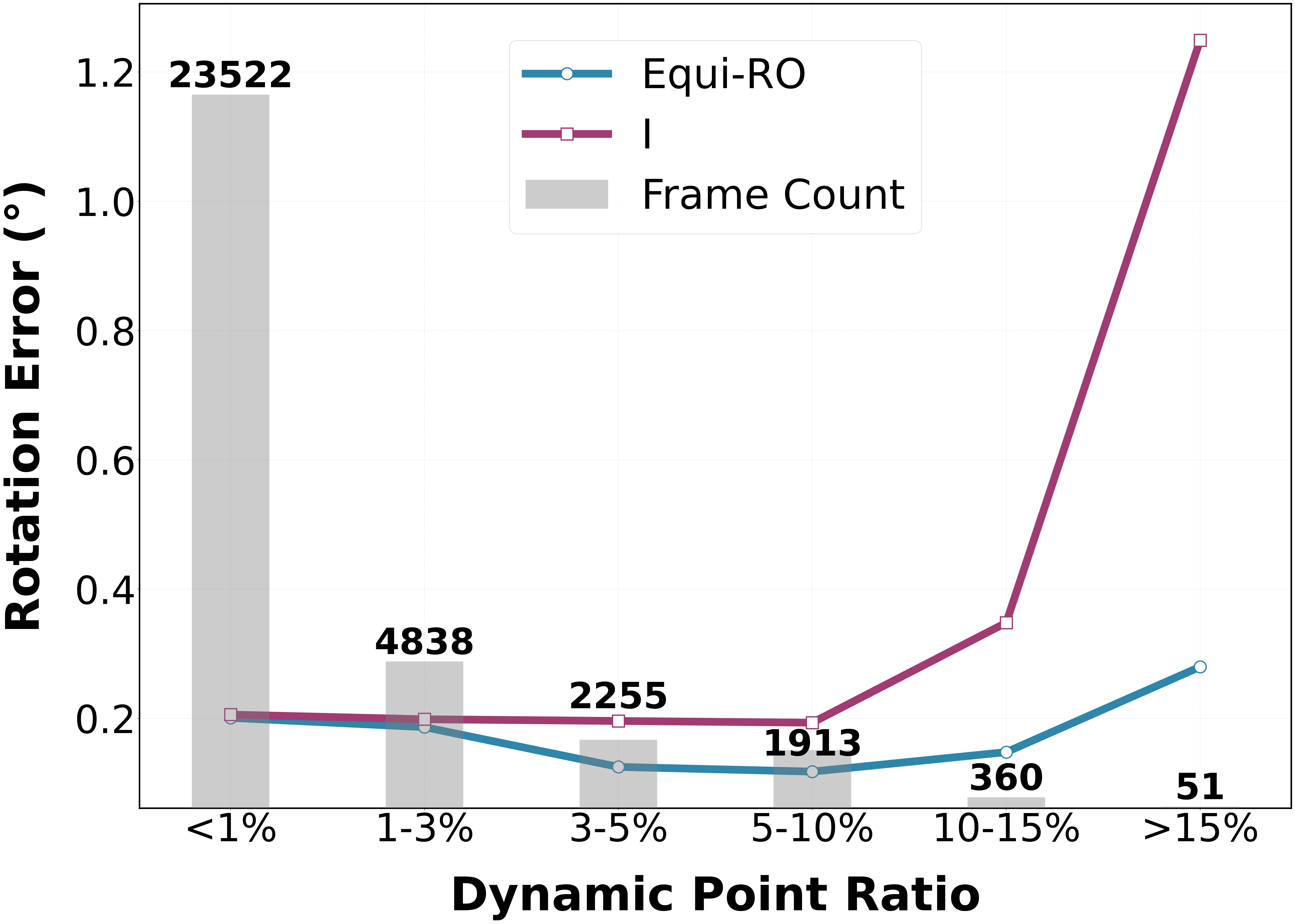}
        \label{Fig:rot_trend}}\subfigure{
        \includegraphics[width=0.23\textwidth]{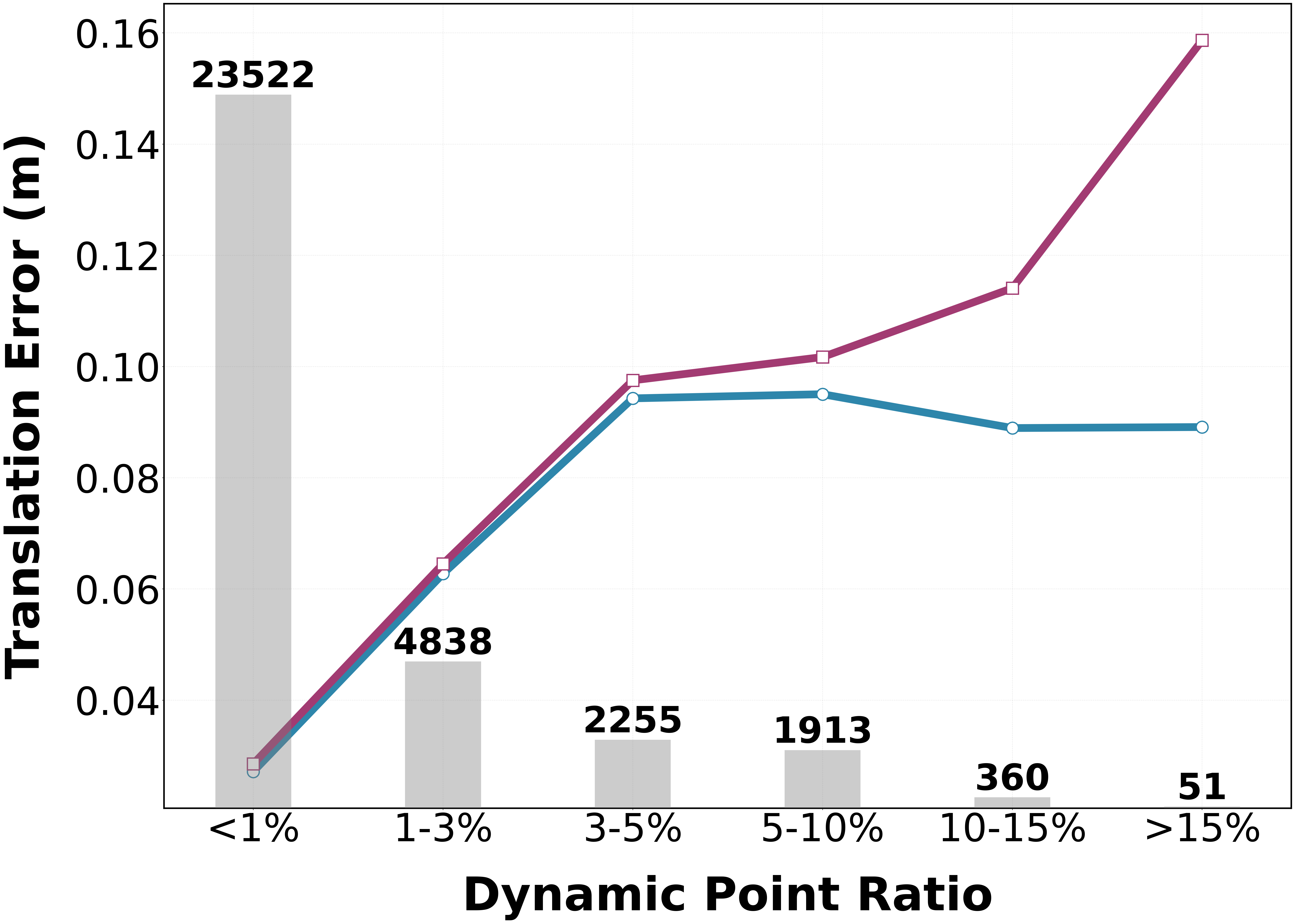}
        \label{Fig:trans_trend}}
    \caption{Rotation (left) and translation (right) errors under different dynamic-point ratios for Equi-RO and the ablation variant I. Points with absolute compensated Doppler velocities greater than 0.1 m/s are treated as dynamic, and numbers on top of bars denote frame counts.}
    \label{Fig:trend}
\vspace{-5mm}
\end{figure}

\subsection{Experimental Setup}

\subsubsection{Training Details}
We train Equi-RO on a single NVIDIA A100 GPU using Adam with a weight decay of $10^{-3}$ and a batch size of 4 for 50 epochs. The initial learning rate is $10^{-4}$ and decays by a factor of 0.1 every 10 epochs. For hyperparameters, we set the weight $\lambda=1$ and the number of nearest neighbors $k=6$ in graph construction.

\subsubsection{Compared Algorithms}
We first select several traditional registration methods ICP, NDT, and GICP as baselines. We also evaluate an ICP-based method KISS-ICP~\cite{vizzo2023kiss}, a 4D radar odometry method Radar4Motion~\cite{kim2024radar4motion}, and Adaptive Probability Distribution-GICP (APDGICP)~\cite{zhang20234dradarslam}, the State-of-the-Art (SOTA) 4D radar odometry method on NTU4DRadLM dataset. For learning-based methods, we include RaFlow~\cite{ding2022self} and CAO-RONet~\cite{li2025cao}. These two methods are also trained on the \textit{loop1} split of NTU4DRadLM using their default settings.

\subsubsection{Metrics}
We adopt the evo library~\cite{grupp2017evo} to compute the relative translation error ($t_{rel}$) and relative rotation error ($r_{rel}$) for each algorithm, by comparing registered trajectories with the ground truth. We average errors over 100 equal segments (with $\delta=L/100$) of each trajectory.

\subsection{Experimental Results}
Fig.~\ref{Fig:quali_vis} qualitatively shows that velocity compensation can help estimate the actual motion of dynamic points, while the feature extraction can select keypoints that remain stationary and consistent between frames.

We present quantitative comparison of our Equi-RO against baseline algorithms in Table~\ref{Tab:results}. In the NTU4DRadLM dataset, our method achieves the lowest translation and rotation errors on the \textit{cp}, \textit{garden}, and \textit{loop3} splits, and ranks second best rotation error on the \textit{nyl} split. GICP, APDGICP, KISS-ICP and Radar4Motion attain competitive results, while learning-based approaches RaFlow and CAO-RONet exhibit inferior performance. This discrepancy is likely because they are originally developed for the earlier VoD dataset~\cite{palffy2022multi}, and appear less suitable to the newer NTU4DRadLM dataset. On average, our algorithm improves translation and rotation accuracy by 10.7\% and 13.4\%, respectively, compared to the best baselines. Fig.~\ref{Fig:vis} visualizes the estimated trajectories, where our method (red lines) demonstrates superior accuracy in most splits. 

The comparative results on self-collected dataset are also shown in Table~\ref{Tab:results} and Fig.~\ref{Fig:self-vis}. All methods degrade on this dataset with lower radar resolution. Nonetheless, our method still attains the best translation and rotation accuracy across both splits, despite being trained solely on the NTU4DRadLM dataset, demonstrating its strong generalization capability.

We further evaluate performance on sharp-turn segments from the NTU4DRadLM dataset. Equi-RO consistently achieves the best or second-best accuracy, as shown in Fig.~\ref{Fig:largeR} and Table~\ref{Tab:largeR}. These results confirm that the proposed equivariant and invariant features, and the equivariant network design are especially effective in scenarios with large rotations. 

\subsection{Ablation Study}
We conduct an ablation study on the NTU4DRadLM dataset by creating five variants of Equi-RO:
\par \noindent \textbf{(I) w/o invariant velocity compensation}: In this variant, node velocities are set to the original Doppler values $v_i^{dop}$, and edge velocities are computed as $v_i^{dop} - v_j^{dop}$, rather than using the compensated counterparts $v_i^{dop'}$ and $v_{ij}$.
\par \noindent \textbf{(II) w/o equivariant network design}: Here, the LN-based layers are replaced by conventional Linear, ReLU, and Pooling layers, and the EGNN is substituted with a standard message passing neural network, removing the equivariance property.
\par \noindent \textbf{(III) w/o adaptive loss}: In this case, the pitch loss $\mathcal{L}_p$ and yaw loss $\mathcal{L}_y$ are omitted, and fixed weights ($s_r = -8.0$, $s_t = -3.0$) are assigned to the rotation and translation losses instead of learnable parameters.
\par \noindent \textbf{(IV) w/o Doppler velocity}: The compensated node velocity $v_i^{dop'}$ and edge velocity $v_{ij}$ are set to zero in this case.
\par \noindent \textbf{(V) w/o initial translation}: In this ablation, we remove the initial translation based on ego-velocity.

The quantitative results are summarized in Table~\ref{Tab:ablation}.

\begin{table}[htbp]
    \centering
    \caption{Ablation study results on the NTU4DRadLM dataset}
    \label{Tab:ablation}
    \resizebox{\linewidth}{!}{
    \begin{tabular}{@{}ccccccccc@{}}
    \toprule
    \multirow{2}{*}{Method} & \multicolumn{2}{c}{cp} & \multicolumn{2}{c}{garden} & \multicolumn{2}{c}{nyl} & \multicolumn{2}{c}{loop3} \\
     \cmidrule(lr){2-3} \cmidrule(lr){4-5} \cmidrule(lr){6-7} \cmidrule(lr){8-9}
     & \( t_{\text{rel}} \) & \( r_{\text{rel}} \) & \( t_{\text{rel}} \) & \( r_{\text{rel}} \) & \( t_{\text{rel}} \) & \( r_{\text{rel}} \) & \( t_{\text{rel}} \) & \( r_{\text{rel}} \) \\
    \midrule
    I    & \underline{5.16}    & 0.0542    & 4.92    & 0.0601    & 8.12& 0.0432& 4.31    & \underline{0.0088}        \\
    II    & 9.60  & 0.1688    & 13.27     & 0.1960    & 8.97     & \underline{0.0376}& 7.97    & 0.0101        \\
    III   & 5.32    & \underline{0.0498}    & 4.76& 0.0744    & 9.11     & 0.0504& 5.24     & 0.0222        \\
    IV   & 5.25    & 0.0523    & 4.94     & 0.0614    & 8.67     & 0.0505& 6.78     & 0.0109        \\
    V   & 5.21    & 0.0514    & \underline{3.97}     & \underline{0.0427}    & \underline{6.87}& 0.0401& \underline{4.05}     & 0.0095        \\
    \textbf{Equi-RO}   & \textbf{3.93} & \textbf{0.0421} & \textbf{3.15} & \textbf{0.0380} & \textbf{4.31}     & \textbf{0.0218} & \textbf{3.25} & \textbf{0.0070}     \\
    \bottomrule
    \end{tabular}}
\end{table}
We also visualize the trajectories generated by each ablated variant across different splits in Fig.~\ref{Fig:ablation}. These trajectories demonstrate that the full Equi-RO model consistently achieves the best performance. All five ablated variants show notable performance degradation. The variant II, which discards equivariant network design, exhibits the worst results across most splits, underscoring the critical importance of equivariant architecture in our framework.

As shown in Fig.~\ref{Fig:trend}, we compare the per-frame average translation and rotation errors of the Equi-RO method with the no invariant velocity compensation variant I across different dynamic-point ratios. As the ratio of dynamic points increases, the errors of the variant I consistently grow, while the full method remains relatively stable. This demonstrates that compensated invariant Doppler features effectively improve robustness in dynamic scenes.

\subsection{Limitations}
Although Equi-RO demonstrates superior accuracy and robustness, its performance is still affected by the relatively low resolution of radars, as shown in the self-collected dataset. Moreover, while our method achieves real-time performance at 58.15 ms per frame on a single NVIDIA A100 GPU, further optimization is necessary for deployment on on-board platforms. Efficiency can be improved via C++ migration and TensorRT acceleration, bridging the gap between the research prototype and mobile platform constraints.

\section{Conclusion}  \label{Sec:Conclusion}

In this paper, we propose a novel 4D mmWave radar odometry algorithm based on equivariant networks. Our method effectively handles noisy, sparse radar point clouds while preserving geometric consistency under large rotations. Extensive experiments on NTU4DRadLM and the self-collected dataset validate its effectiveness and generalizability. This work highlights the potential of equivariant network design for radar-based perception and provides a new perspective for point cloud processing in robotics and autonomous systems. Future work will extend the framework to multi-modal fusion with IMUs and cameras, and explore its application to other 3D point clouds with physical attributes.

\FloatBarrier
\bibliographystyle{IEEEtran} 
\bibliography{strings-abrv,ieee-abrv,ref}

\end{document}